\documentclass[10pt,twocolumn,letterpaper]{article}

\usepackage{cvpr}              
\usepackage{algorithm}
\usepackage{algpseudocode}
\usepackage{siunitx}
\usepackage{adjustbox}
\usepackage{anyfontsize}
\usepackage[normalem]{ulem}
\robustify\bfseries
\robustify\uline
\def\Uline#1{#1\llap{\uline{\phantom{#1}}}}
\newrobustcmd\B{\DeclareFontSeriesDefault[rm]{bf}{b}\bfseries}                          %

\usepackage{multirow}
\usepackage{amsmath}
\DeclareMathOperator*{\argmax}{argmax}

\definecolor{cvprblue}{rgb}{0.21,0.49,0.74}

\usepackage[pagebackref,breaklinks,colorlinks,allcolors=cvprblue]{hyperref}

\def\paperID{13307} 
\def\confName{CVPR}
\def\confYear{2025}

\title{Benchmarking Object Detectors under Real-World Distribution Shifts in Satellite Imagery}

\author{
    Sara A. Al-Emadi$^{1,2}$ \quad
    Yin Yang$^{2}$ \quad
    Ferda Ofli$^{1}$  \\
    $^1$ Qatar Computing Research Institute, HBKU \&
    $^2$ College of Science and Engineering, HBKU \\
    {\tt\small \{salemadi, yyang, fofli\}@hbku.edu.qa}
}

\begin{document}
\maketitle
\begin{abstract}
Object detectors have achieved remarkable performance in many applications; however, these deep learning models are typically designed under the i.i.d. assumption, meaning they are trained and evaluated on data sampled from the same (source) distribution. In real-world deployment, however, target distributions often differ from source data, leading to substantial performance degradation. Domain Generalisation (DG) seeks to bridge this gap by enabling models to generalise to Out-Of-Distribution (OOD) data without access to target distributions during training, enhancing robustness to unseen conditions. In this work, we examine the generalisability and robustness of state-of-the-art object detectors under real-world distribution shifts, focusing particularly on spatial domain shifts. Despite the need, a standardised benchmark dataset specifically designed for assessing object detection under realistic DG scenarios is currently lacking. To address this, we introduce Real-World Distribution Shifts (RWDS), a suite of three novel DG benchmarking datasets that focus on humanitarian and climate change applications. These datasets enable the investigation of domain shifts across (i) climate zones and (ii) various disasters and geographic regions. To our knowledge, these are the first DG benchmarking datasets tailored for object detection in real-world, high-impact contexts. We aim for these datasets to serve as valuable resources for evaluating the robustness and generalisation of future object detection models. Our datasets and code are available at \url{https://github.com/RWGAI/RWDS}.
\end{abstract}

\vspace{-2mm}    
\section{Introduction}
\label{sec:intro}
Deep learning has achieved remarkable success in various applications, including flood mapping~\cite{chp1_app_floodmapping_1, ferda_flood_mapping_rs_app}, medical diagnostics~\cite{chp1_app_medicalimaging_1, chp1_app_medicalimaging_2}, and self-driving cars~\cite{chp1_app_selfdriving_1, chp1_app_selfdriving_2}. However, these machine learning models are typically developed under the i.i.d. assumption, where they are trained and evaluated on data samples drawn from the same \textit{source} distribution.
Consequently, when deployed in real-world environments with differing \textit{target} distributions, these models experience significant performance degradation, hindering their large-scale deployment. This phenomenon is known as distribution or domain shift. In this paper, we focus a specific type of domain shift, referred to as spatial domain shift (i.e., covariate shift) on a global scale, which is driven by visual variations in land cover and built structures influenced by factors such as natural landscapes, climate zones, architectural styles, economic and financial development, social and cultural attributes, and human settlement patterns.

\begin{figure}[t]
\centering
\small
\centerline{\includegraphics[height=3.5cm,keepaspectratio]{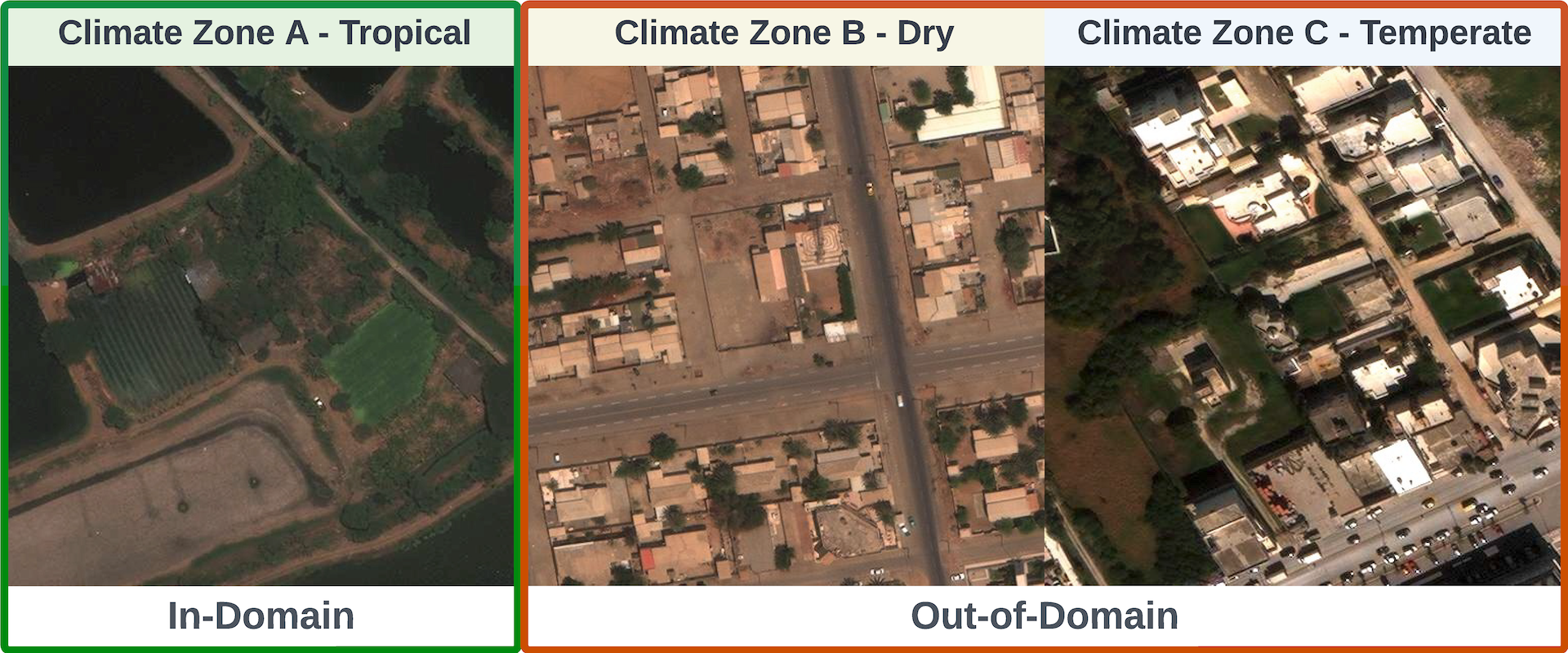}}
\caption{Example images from different climate zones}
\label{fig:fig_1_climate_zones_representation}
\vspace{-4mm}
\end{figure}

In satellite imagery-based object detection, spatial domain shift poses a significant challenge, especially when environmental conditions vary unpredictably, as demonstrated in Figure~\ref{fig:fig_1_climate_zones_representation}. To examine this, we use the Köppen climate classification system~\cite{koppen_classification_wiki_2024} and consider a scenario where an object detector is trained on images from a \textit{tropical} climate zone but evaluated on \textit{Out-of-Distribution (OOD)}  target domains, specifically \textit{dry} and \textit{temperate} climate zones. Since these target domains exhibit distinct visual characteristics compared to the source domain, a performance drop is expected, highlighting the impact of spatial domain shift when applying object detection models across diverse climatic contexts.

Several studies have sought to mitigate the issue of domain shift through data augmentation~\cite{data_aug_paper}, transfer learning~\cite{TL_in_RS_RSE}, and domain adaptation, in which the model has access to unlabelled samples from the target distribution during training~\cite{lr_p_9,lr_p_10}. However, in real-world applications, models often encounter distributions that cannot be foreseen before deployment. To tackle this challenge, recent research has focused on domain shift under this constraint, a problem known as Domain Generalisation (DG).

DG datasets are essential for assessing a model's ability to generalise to unseen target distributions. In image classification, a substantial body of research have been devoted to curating DG datasets for broad use cases, such as PACS~\cite{pacs} and DomainNet~\cite{domainnet}, as well as datasets introducing synthetic domain shifts, like RotatedMNIST~\cite{rotatedMNIST}, or real-world distribution shifts, as seen in WILDS~\cite{wilds}. However, the study of domain shift in object detection remains relatively underexplored. To bridge this gap, Mao \etal introduced COCO-O~\cite{coco-o}, a DG benchmark for object detection with six domains including Sketch, Weather, Cartoon, Painting, Tattoo, and Handmade, to evaluate both in-domain (ID) and OOD performance. While the domain shifts in COCO-O are evident, such as the differences between sketches and paintings, further investigation of the practical motivation for training a model on sketches and testing it on paintings could provide valuable insights. To our knowledge, a DG benchmark does not currently exist for evaluating the behaviours of object detectors on OOD test data in a common, real-world application setting.

Motivated by this, we introduce \textit{Real-World Distribution Shifts} (\textit{RWDS}), a suite of three realistic DG benchmarking datasets, namely, RWDS-CZ, RWDS-FR and RWDS-HE, which focus on humanitarian and climate change applications and investigate domain shifts across (i) climate zones and (ii) different disasters and geographic regions, respectively. Moreover, we benchmark and analyse the performance of several state-of-the-art (SOTA) object detection algorithms on RWDS under two setups: single-source, where an object detector is trained on only one source domain, and multi-source, where training incorporates multiple source domains. We then evaluate these models on the unseen target domains to provide comprehensive insights into their generalisation performance. We trained around 100 object detector models and conducted over 200 experiments. Our contributions are summarised as follows:

\begin{itemize}
    \item We propose RWDS, a suite of novel, realistic and challenging DG datasets designed to evaluate spatial domain shifts in real-world object detection tasks.

    \item We provide the community with in-depth benchmarking analyses on the performance of the SOTA object detectors on RWDS datasets.

    \item We analyse the impact of single-source versus multi-source training in DG for spatial domain shifts in satellite imagery, concluding that multi-source training enhances generalisability of object detectors.

\end{itemize}

The rest of the paper is organised as follows: Section~\ref{sec:related_work} reviews literature on object detection and robustness benchmarks. Section~\ref{sec:Our_RWDS_Dataset} introduces the RWDS datasets with details on data cleaning and preprocessing. Section~\ref{sec:Experiments} describes the evaluation metrics, experimental setups, and selected object detectors. Section~\ref{sec:Results_and_Analyses} presents the results and provides a comprehensive analysis and Section~\ref{sec:Conclusion} concludes the paper.

\section{Related Work}
\label{sec:related_work}

\subsection{Object Detection}
Early attempts in deep-learning-based object detection used a set of bounding boxes and masked regions as input to the CNN architecture to incorporate shape information into the classification process to perform object localisation~\cite{girshick2014rcnn, szegedy2013deep, Erhan:CVPR14, Szegedy:arXiv15v3}. Later on, end-to-end techniques were proposed based on shared computation of convolutions for simultaneous detection and localization of the objects~\cite{Hariharan:ECCV14, Girshick:2015vr, sermanet-iclr-14, Dai:CVPR15, ren15gu, Liu2016ssd, redmon2016you, he2017mask}. These methods can be generally divided into two categories: one-stage detectors~\cite{redmon2016you, Liu2016ssd, ross2017focal, redmon2017yolo9000, law2018cornernet, tian2019fcos, duan2019centernet, zhou2019bottom, tan2020efficientdet, tood_paper} and two-stage detectors~\cite{girshick2014rcnn, Girshick:2015vr, he2015spatial, faster_rcnn_paper, he2017mask, lin2017feature, cai2019cascade, sun2021sparse}. More recently, transformer-based object detection models have proved more efficient and accurate, thanks to their ability to not require anchor boxes and non-maximum suppression procedure~\cite{carion2020end, zhu2021deformable, dino_paper}. Besides, with the advances in foundation models (large vision models or vision-language models), open-set and open-world object detection has become popular~\cite{grounding_dino_paper, glip_paper, wu2024grit}. Following these trends, remote sensing community has also integrated deep learning-based object detection models into their research~\cite{long2017accurate, liu2018detection, yang2019scrdet, li2020object, ding2021object, han2021redet, sahi}. However, accurate object detection from satellite imagery \emph{at scale} remains a challenging task.

\subsection{Robustness Benchmarks}
Various benchmark studies have been developed to assess the robustness of object detection models under distribution shifts. For instance, COCO-C~\cite{michaelis2019benchmarking} evaluates model performance by applying synthetic corruptions, such as JPEG compression and Gaussian noise, to the COCO test set. Similarly, OOD-CV~\cite{zhao2022ood} and its extended version, OOD-CV-v2~\cite{zhao2024ood}, include OOD examples across 10 object categories from PASCAL VOC and ImageNet, spanning variations in pose, shape, texture, context, and weather conditions. These datasets enable benchmarking across multiple tasks like image classification, object detection, and 3D pose estimation. COCO-O~\cite{coco-o} introduces natural distribution shifts in COCO-based datasets, spanning six domains—weather, painting, handmade, cartoon, tattoo, and sketch. Their study has shown that there is a significant performance gap of 55.7\% between ID and OOD performance, highlighting the domain generalisation challenges under such shifts. However, despite their contributions, these datasets still lack the complexity of real-world distribution shifts. More realistic benchmarks include those reflecting environmental changes in autonomous driving~\cite{johnson2017driving} and object variations in aerial imagery~\cite{xia2018dota}, which better capture the dynamic and unpredictable conditions faced in practical applications. However, they remain limited in scope, as they do not comprehensively account for geographic and temporal variability, environmental and weather conditions, occlusion, clutter, and object appearance changes within a unified framework. In contrast, our RWDS datasets aim to bridge this gap by providing a diverse and realistic evaluation setting that encapsulates these real-world domain shifts more holistically.
\begin{figure*}[t] 
\centering
\subfloat[Climate Zone A]{\includegraphics[width=0.333\linewidth]{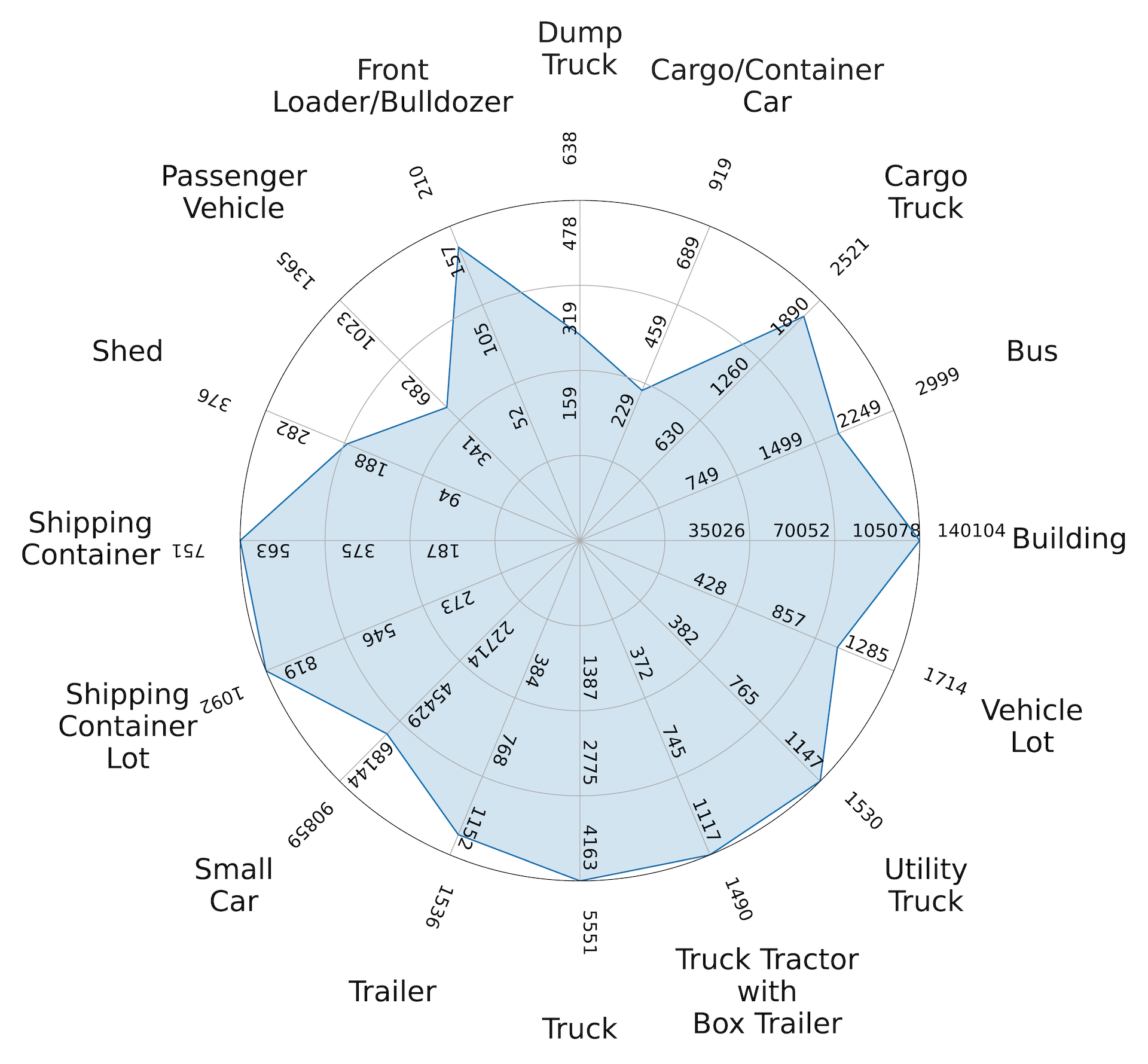}}\hfil
\subfloat[Climate Zone B]{\includegraphics[width=0.333\linewidth]{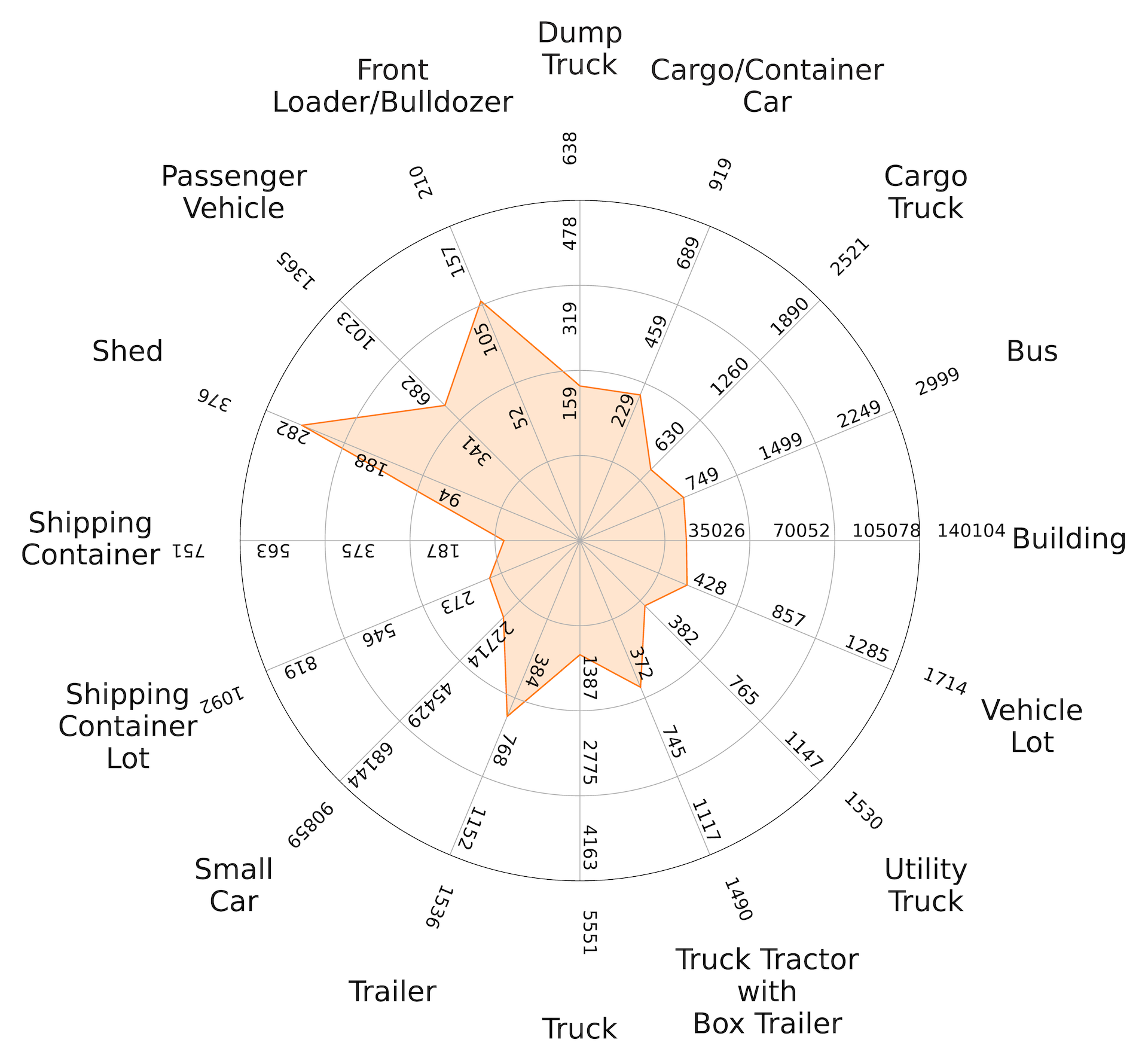}}\hfil 
\subfloat[Climate Zone C]{\includegraphics[width=0.333\linewidth]{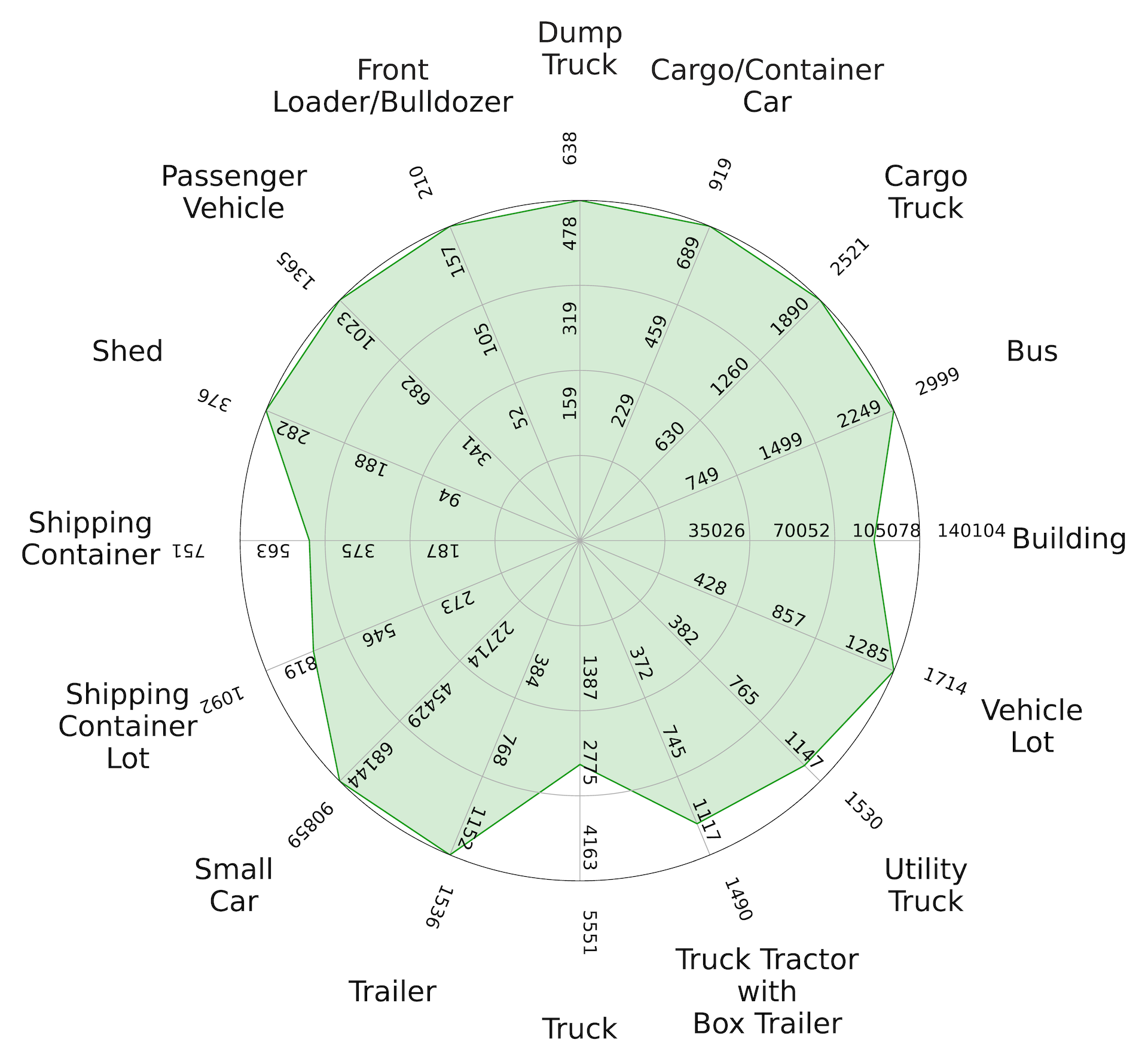}} 
\caption{Class-wise distribution of training data for each domain as well as the overall data distribution across domains in RWDS-CZ}\label{fig:rwds_cz_dataset_distribution}
\end{figure*}

\section{Our RWDS Datasets}\label{sec:Our_RWDS_Dataset}
\subsection{RWDS Across Climate Zones}\label{ssec:Our_RWDS_CZ_Dataset}
While there are increasing efforts to mitigate and reduce the negative and potential impact of climate change on the global ecosystem including natural resources, weather and the natural landscapes~\cite{climate_change_paper}, there is a need to develop robust models to support computer vision tasks under these circumstances, more particularly, object detection task. In order to investigate their robustness and generalisability across different climate zones, we propose \textbf{RWDS across Climate Zones (RWDS-CZ)} dataset where we focus on Köppen's climate zone classification~\cite{koppen_classification_wiki_2024, koppen_classification_national_geographic_article}. Given the scarcity of global satellite imagery that covers all climate zones, we use the raw satellite imagery from the xView dataset~\cite{xview_dataset}, an open-source object detection dataset featuring high-resolution (0.3m) images captured at a global scale across 60 object classes. For this study, we focus on three distinct climate zones: Zone A (CZ~A)—tropical or equatorial, Zone B (CZ~B)—arid or dry, and Zone C (CZ~C)—warm/mild temperate. These serve as our distinct domains for studying spatial domain shifts in satellite-based object detection.

To create the domains, we first map the geo-coordinates of each image to its respective climate zone and proceed with splitting the overall dataset into domains. However, this results in a mismatch between the classes available across the domains. To resolve this, we retain only those classes that appear in all climate zones. Additionally, we set a threshold of 30 samples per class to ensure sufficient data for training. Any class with fewer than 30 samples is excluded from all domains.
This process yields a total of 16 classes. To maintain consistent distribution of object instances across the training, validation, and test sets within each domain, we follow the procedure outlined in Algorithm~\ref{alg:dsp}. This process is repeated for each domain, resulting in the final RWDS-CZ dataset. Table~\ref{tab:rwds_cz_per_set_dist} summarises the dataset statistics, while Figure~\ref{fig:rwds_cz_dataset_distribution} shows the distribution of training samples across classes in all domains.

\begin{algorithm}[t]
\small
\caption{Dataset Split Procedure}\label{alg:dsp}
\begin{algorithmic}[1]
\State \textbf{Input:} Set of images $\mathcal{I}$, class labels $\mathcal{C}$
\State \textbf{Initialise:} Training set $\mathcal{T} \gets \emptyset$, Validation set $\mathcal{V} \gets \emptyset$, Testing set $\mathcal{S} \gets \emptyset$
\Function{AllocImages}{$\mathcal{I},\mathcal{N}$}

        \For{each class $c \in \mathcal{C}$}
            \State $I^* \gets \argmax\limits_{I \in \mathcal{I}} \text{count}(I, c)$ \Comment{Select image with most instances of $c$}
            \State $\mathcal{N} \gets \mathcal{N} \cup \{I^*\}$ \Comment{Append to designated set}
            \State $\mathcal{I} \gets \mathcal{I} \setminus \{I^*\}$ \Comment{Remove allocated image}
        \EndFor
    \State \Return $(\mathcal{N}, \mathcal{I})$ \Comment{Return final dataset splits}

    \EndFunction
    \While{$\mathcal{I} \neq \emptyset$}
        \For{$i = 1$ to $3$} \Comment{Repeat 3 times for training set}
            \State $\mathcal{T},\mathcal{I} \gets$ \Call{AllocImages} {$\mathcal{I}, \mathcal{T}$} \Comment{Update training set}
        \EndFor
        \State $\mathcal{V},\mathcal{I} \gets$ \Call{AllocImages} {$\mathcal{I},\mathcal{V}$} \Comment{Update validation set}
        \State $\mathcal{S},\mathcal{I}  \gets$ \Call{AllocImages}{$\mathcal{I},\mathcal{S}$} \Comment{Update test set}
    \EndWhile
\end{algorithmic}
\end{algorithm}

\begin{table}[t]
\centering
\renewcommand{\arraystretch}{1.1}
\fontsize{9pt}{9pt}\selectfont
\setlength{\tabcolsep}{18pt}
\begin{tabular}{@{}lrrr@{}}
\toprule
Split   &\multicolumn{1}{c}{CZ A}& \multicolumn{1}{c}{CZ B} & \multicolumn{1}{c@{}}{CZ C}   \\ \midrule
Training & $117,265$ & $43,272$ & $124,717$\\
Validation  & $58,997$ & $13,423$ & $47,362$\\
Test  & $56,954$ & $24,745$ & $60,310$ \\ \bottomrule
\end{tabular}
\caption{RWDS-CZ overall object instances per partition}
\label{tab:rwds_cz_per_set_dist}
\end{table}

\begin{figure}[t]
\centering
\small
\centerline{\includegraphics[width=\columnwidth,keepaspectratio]{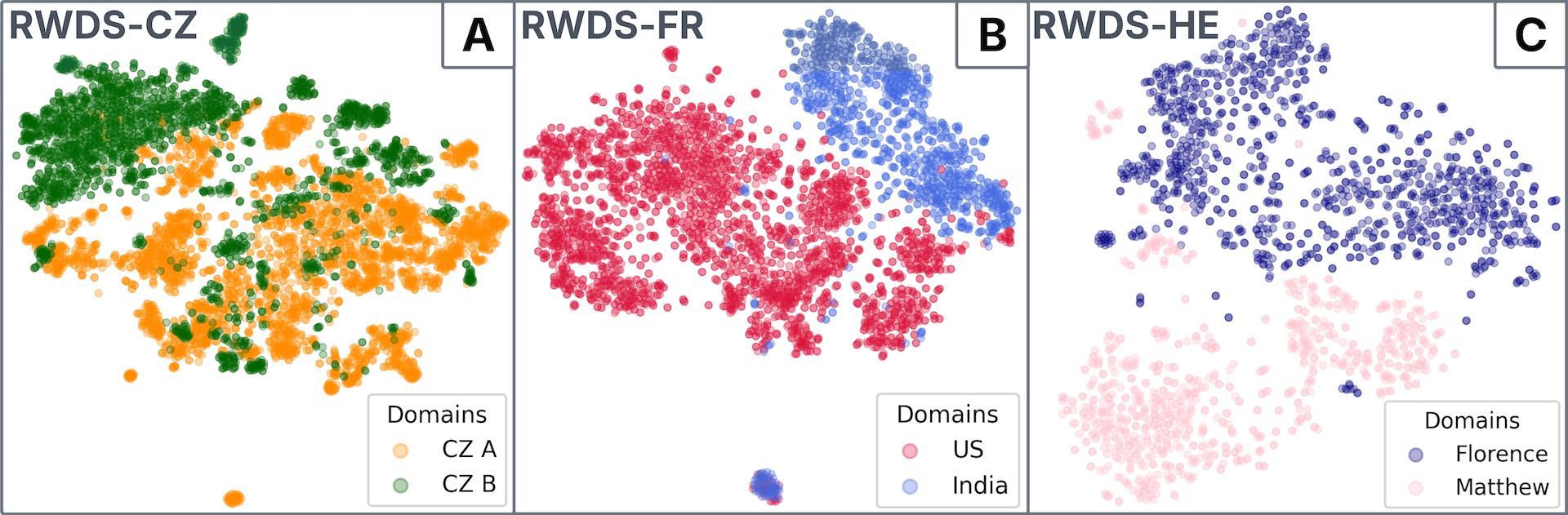}}
\caption{Embedding space representations of the RWDS datasets}
\label{fig:rwds_tsne_representation}
\vspace{-3mm}
\end{figure}

To visualize the domain shift in RWDS-CZ, we extract image embeddings using RemoteCLIP~\cite{mllm_remoteclip} and project them into 2D using t-SNE~\cite{tsne}. Figure~\ref{fig:rwds_tsne_representation}-A showcases the shift between images from CZ~A and CZ~B.

\subsection{RWDS in Disaster Damage Assessment}

A notable consequence of climate change is the increasing frequency and severity of natural disasters such as hurricanes, storms, floods, wildfires, earthquakes, tsunamis, etc. Damage assessment is essential during and after disasters to support aid delivery, guide building reconstruction efforts, and provide governments and humanitarian agencies with an estimate of the disaster's impact. Generally, large amount of satellite imagery is captured around the disaster-hit locations. However, given the sheer volume of data, the cleaning, preprocessing and re-training of object detectors for each disaster on the spot is time consuming and might not be feasible due to lack of annotations, highlighting the crucial need for robust models that can generalise well to unseen distributions beyond those they were trained on. Hence, we investigate the robustness of SOTA object detectors in this application under two different scenarios.

In the first use-case, we examine the shift of the same disaster type across distant geographic regions with different socioeconomical characteristics. Specifically, we define domains in terms of collection of events that caused \textit{floods} across the United States and India, respectively. We refer to this use-case as \textbf{RWDS across Flooded Regions (RWDS-FR)}. Whereas, in the second use-case, we focus on understanding the shift in the behaviour of these models across different disaster events of the same type, namely, \textit{hurricanes}, in North America. We refer to this use-case as \textbf{RWDS across Hurricane Events (RWDS-HE)}. Similar to the discussion related to the scarcity of open-source satellite imagery, we utilise the raw satellite images released in the xDB building damage assessment dataset~\cite{xview2_dataset} for both RWDS-FR and RWDS-HE.

\subsubsection{RWDS Across Flooded Regions (RWDS-FR)}\label{sec:dataset_rwds_fr}
\begin{figure}[t]
\centering
\small
\centerline{\includegraphics[height=3cm,keepaspectratio]{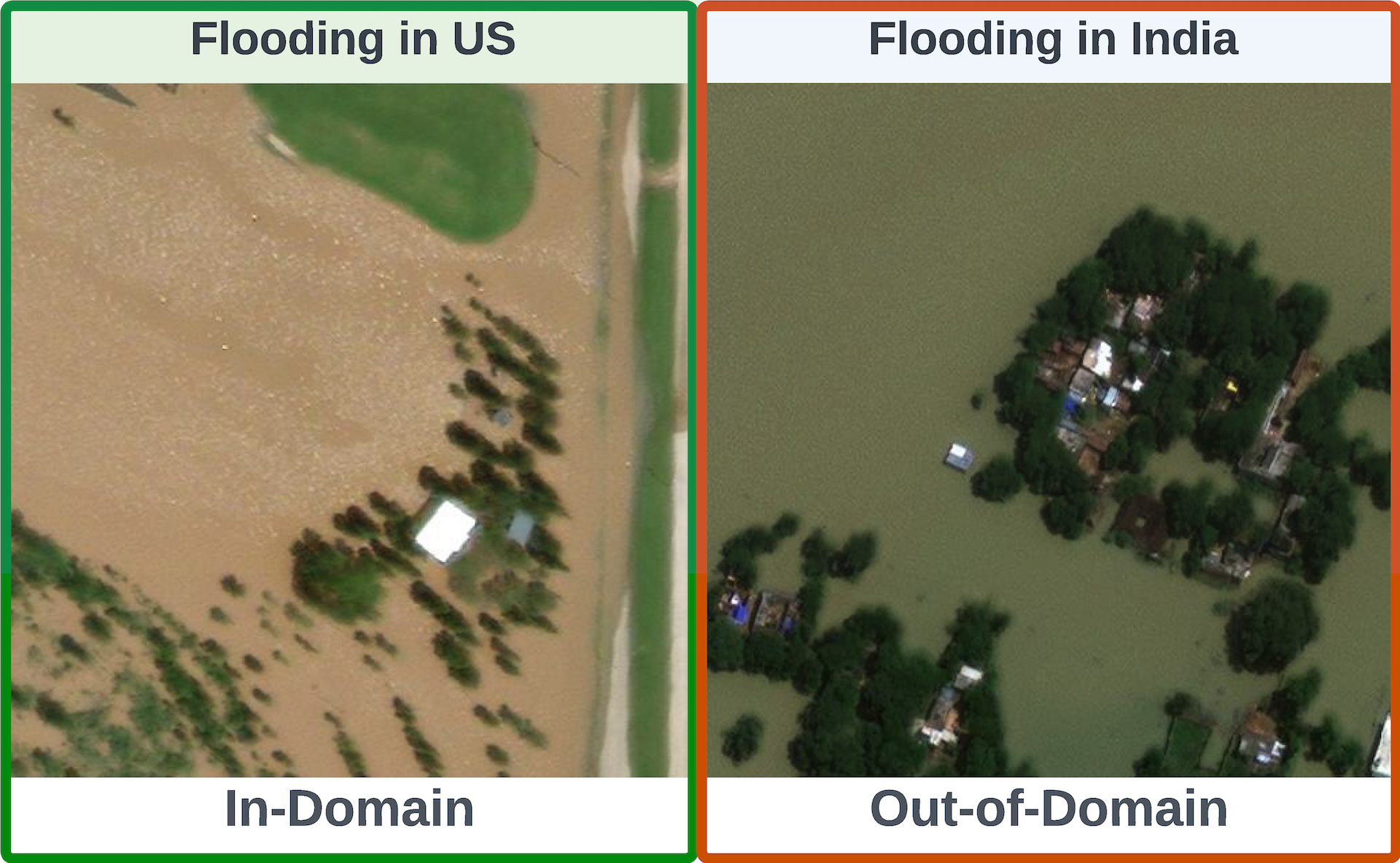}}
\caption{Comparison of flood scenes between the US and India}
\label{fig:rwds_fr_diagram_v2}
\end{figure}

We start by creating the metadata for the raw images. xDB dataset provides disaster event, damage type, and polygons of buildings for segmentation application. However, given that we are interested in object detection, we convert polygons of buildings into bounding boxes. Furthermore, similar to Section~\ref{ssec:Our_RWDS_CZ_Dataset}, we map the latitude and longitude coordinates of the polygons to find the corresponding location of each object instance in terms of country, region, continent, etc. We then extract the flooded objects in India and US. Figure~\ref{fig:rwds_fr_diagram_v2} shows example images illustrating the shift between the domains, with a close-up visualization of image embeddings in Figure~\ref{fig:rwds_tsne_representation}-B. Unlike the original data, where the instances are categorised into four classes, namely, no damage, minor damage, major damage and destroyed, when extracting the flooded instances in the US and India, we observe a class imbalance between those classes. Therefore, we transform the task into a binary categorisation, leaving us with two classes, namely, damaged (D) and no damage (ND). We then follow the same logic as that discussed in Section~\ref{ssec:Our_RWDS_CZ_Dataset} to create the training, validation and testing splits. This yields the RWDS-FR dataset. Table~\ref{tab:RWDS_FD_dataset_dist_overall} represents the resulting domain and class distributions per split.

\begin{table}[t]
\centering
\renewcommand{\arraystretch}{1.1}
\fontsize{9pt}{9pt}\selectfont
\setlength{\tabcolsep}{12pt}
\begin{tabular}{@{}lrrrr@{}}
\toprule
 & \multicolumn{2}{c}{India} & \multicolumn{2}{c}{US}\\
 \cmidrule(lr){2-3}\cmidrule(lr){4-5}

Split   &\multicolumn{1}{c}{D}& \multicolumn{1}{c}{ND} &\multicolumn{1}{c}{D}& \multicolumn{1}{c@{}}{ND} \\ \midrule
Training &  $5,023$ &  $14,841$  &  $10,680$  &  $20,055$  \\
Validation  & $2,532$  & $8,320$  &  $5,470$  &  $9,834$ \\
Test  &  $2,802$  &  $8,064$  &  $5,452$  &  $10,034$ \\ \bottomrule
\end{tabular}
\caption{RWDS-FR object instances per partition.}
\label{tab:RWDS_FD_dataset_dist_overall}
\vspace{-3mm}
\end{table}

\subsubsection{RWDS Across Hurricane Events (RWDS-HE)}\label{sec:D_rwds_HE}

\begin{figure}[t]
\centering
\small
\centerline{\includegraphics[height=2.6cm,keepaspectratio]{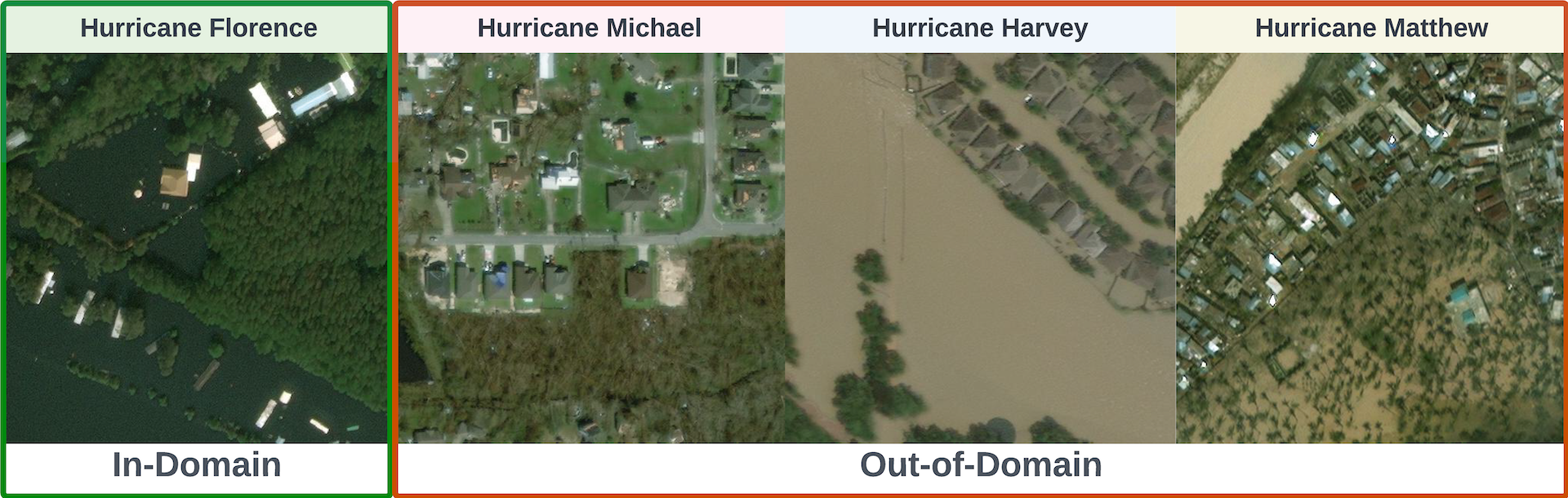}}
\caption{Comparison of hurricane scenes from different events}
\label{fig:rwds_he_diagram}
\end{figure}

In contrast to RWDS-FR dataset, where domains are defined by diverse geographic regions, this dataset focuses on hurricane events across North America, which are geographically closer in proximity. As a result, the dataset consists of four hurricane events as domains: Florence, Michael, Harvey, and Matthew, as shown in Figure~\ref{fig:rwds_he_diagram}. Figure~\ref{fig:rwds_tsne_representation}-C presents the image embeddings for hurricanes Florence and Matthew as an example. We adhere to the preprocessing, metadata creation, binary class categorisation, and data splitting procedures outlined in Section~\ref{sec:dataset_rwds_fr}. This defines the RWDS-HE dataset. Table~\ref{tab:RWDS_HE_dataset_dist_overall} presents the final per-class and per-split distribution for each domain.  

\begin{table}[t]
\centering
\renewcommand{\arraystretch}{1.2}
\setlength{\tabcolsep}{3pt}

\begin{adjustbox}{max width=\columnwidth}
\fontsize{8.85pt}{8.85pt}\selectfont

\sisetup{
  table-format=4.0,
  table-number-alignment=center,
  table-text-alignment=center,
  detect-weight        = true,
  input-decimal-markers = .,input-ignore = {,}, group-separator={,},
  group-four-digits = true
}

\begin{tabular}{@{}lSSSSSSSS@{}}
\toprule
 & \multicolumn{2}{c}{Florence} & \multicolumn{2}{c}{Michael} & \multicolumn{2}{c}{Harvey}& \multicolumn{2}{c}{Matthew}\\
 \cmidrule(lr){2-3}\cmidrule(lr){4-5}\cmidrule(lr){6-7}\cmidrule(lr){8-9}
Split   &\multicolumn{1}{c}{D}& \multicolumn{1}{c}{ND} &\multicolumn{1}{c}{D}& \multicolumn{1}{c}{ND} &\multicolumn{1}{c}{D}& \multicolumn{1}{c}{ND} &\multicolumn{1}{c}{D}& \multicolumn{1}{c}{ND} \\ \midrule
Training &  1,102 &  4,196 &  6,132 & 11,347 & 9,270 & 9,223 &  8,919 &  1,938 \\
Validation  & 578  &  2,112 & 3,075 &  5,455 & 4,670 & 4,594 & 4,910 & 1,042 \\
Test  & 582 &  2,158 &  3,229  & 5,890  & 4,796 & 4,821 & 4,743 & 1,078 \\ \bottomrule
\end{tabular}
\end{adjustbox}
\caption{RWDS-HE object instances per partition.}
\label{tab:RWDS_HE_dataset_dist_overall}
\vspace{-3mm}
\end{table}
\section{Experiments}\label{sec:Experiments}

\subsection{Single-Source and Multi-Source Setup}\label{sec:E_SS_MS}
We investigate DG in two setups. The first involves training an object detector on a single ID source training set, then assessing its performance on both the held-out OOD target domains and the ID test set. This setup reflects scenarios with a limited diversity of data distributions. Whereas, in the second setup, we incorporate training an object detector on a collection of source domains, mirroring real-world scenarios, where data from a variety of distributions may be available. For quantitative comparison, we evaluate the trained object detector on each OOD target domain separately, as well as on the average performance across ID domains.\footnote{The single- and multi-source setups are formally defined in Supplementary~\ref{sup_sec:maths_formulation_dg_setup}.}

\subsection{DG Evaluation Metrics}
Methods for evaluating DG models remain an active and open area of investigation. Researchers have, however, adapted existing approaches to assess the performance of deep learning models on OOD datasets for classification tasks. Among these, the \textit{leave-one-domain-out} evaluation strategy~\cite{domainbed} is widely regarded. In this setup, one domain is excluded from training, enabling it to serve as an independent test domain to rigorously evaluate model performance without any additional tuning. Inspired by this, we adapt this evaluation technique for object detection under the single- and multi-source setups.

We assess the performance of the object detectors using the standard mean Average Precision (mAP) metric which is commonly used in object detection applications~\cite{survery_od_20years}. More specifically, we use the MS-COCO AP metric, which is calculated as the average over multiple IoU thresholds ranging from 0.50 to 0.95 with a stepsize of 0.05.\footnote{$\text{mAP}_{\text{50}}$ \& $\text{mAP}_{\text{75}}$ results are included in Supplementary~\ref{sup_sec:detailed_benchmarking_results}.} 

\smallskip
\noindent\textbf{Performance Drop (PD).} A metric frequently used in the DG community for assessing the generalisability of classification tasks is the \textit{Performance Drop}, which quantifies the percentage of performance degradation observed in the model when subjected to OOD data from a target domain. Drawing inspiration from this approach, we apply it in the context of DG for object detection. This is formulated as follows:
\begin{equation}\label{eqn:2}
PD = 100\times{\frac{mAP_{ID}-mAP_{OOD}}{mAP_{ID}}}
\end{equation}
where $mAP_{ID}$ and $mAP_{OOD}$ represent the average mAP of the combination of detectors tested on a specific domain's ID and OOD test sets, respectively.

\smallskip
\noindent\textbf{Harmonic Mean (H).} 
To compare the ID and OOD performance of object detectors based on their mAP, we adopt the widely recognised \textit{Harmonic Mean} as another evaluation metric. This choice is motivated by its use in recent generalised open-set zero-shot learning studies~\cite{xian2017zero, fu2019vocabulary, chen2020boundary} to compute a joint score reflecting model performance across in-domain and out-of-domain test sets. This is defined as:

\begin{equation}\label{eqn:3}
H = \frac{2\times{mAP_{OOD}\times{mAP_{ID}}}}{{mAP_{OOD}}+{mAP_{ID}}}
\end{equation}

\begin{table}[t]
\centering
\renewcommand{\arraystretch}{1.1}
\fontsize{9pt}{9pt}\selectfont
\begin{tabular}{@{}ll@{}}
\toprule
\multicolumn{1}{@{}l}{Object Detector}& \multicolumn{1}{l}{Backbone} \\ \midrule
Faster R-CNN~\cite{faster_rcnn_paper} & ResNeXt-101-64x4d \\
Mask R-CNN~\cite{he2017mask}  & ResNeXt-101-64x4d and FPN \\
TOOD~\cite{tood_paper}  & ResNeXt-101-64x4d, DCNv2 and FPN \\ 
DINO (5scale)~\cite{dino_paper}  & Swin-L\\ 
Grounding DINO~\cite{grounding_dino_paper}  & BERT and Swin-B\\ 
GLIP (L)~\cite{glip_paper}  & BERT and Swin-L \\
\bottomrule
\end{tabular}
\caption{Object detectors and their backbone architectures.}
\label{tab:od_arch_bb}
\vspace{-2mm}
\end{table}

\subsection{Object Detectors and Hyperparameters}\label{sec:od_hyp}

\begin{table*}[t]
\centering
\fontsize{9pt}{9pt}\selectfont
\renewcommand{\arraystretch}{1.1}
\setlength{\tabcolsep}{6.3pt}
\sisetup{
  table-format=2.1,
  table-number-alignment=center,
  table-text-alignment=center,
  detect-weight        = true,
}
\begin{tabular}{@{}l
                S[table-format=2.1]
                S[table-format=1.1]
                S[table-format=2.0]
                S[table-format=1.1]
                S[table-format=2.1]
                S[table-format=2.1]
                S[table-format=2.0]
                S[table-format=2.1]
                S[table-format=2.1]
                S[table-format=1.1]
                S[table-format=2.0]
                S[table-format=1.1]@{}}

\toprule
 & \multicolumn{12}{c}{Target} \\
\cmidrule(lr){2-13}
 & \multicolumn{4}{c}{CZ A} & \multicolumn{4}{c}{CZ B} & \multicolumn{4}{c}{CZ C} \\
 \cmidrule(lr){2-5}\cmidrule(lr){6-9}\cmidrule(lr){10-13}
Methods & \multicolumn{1}{c}{$\text{mAP}_\text{ID}$} & $\text{mAP}_\text{OOD}$ & $\text{PD}\downarrow$  & \multicolumn{1}{c}{$\text{H}\uparrow$ } &$\text{mAP}_\text{ID}$ & $\text{mAP}_\text{OOD}$ & $\text{PD}\downarrow$  & \multicolumn{1}{c}{$\text{H}\uparrow$ } & \multicolumn{1}{c}{$\text{mAP}_\text{ID}$} & $\text{mAP}_\text{OOD}$ & $\text{PD}\downarrow$& \multicolumn{1}{c}{$\text{H}\uparrow$ } \\
\midrule
Faster R-CNN &  7.2 &	 3.9 &	 47 &	 5.0 &	 7.5 &	 6.0 &	 20 &	 6.7 &	 7.7 &	 3.4 &	 56 &	 4.7 \\

Mask R-CNN &  7.3 &	 3.7 &	 49 &	 4.9 &	 7.7 &	 5.8 &	 25 &	 6.6 &	 7.8 &	 3.5 &	 55 &	 4.8 \\

TOOD &  7.8 &	 4.0 &	 49 &	 5.2 &	 7.8 &	 6.1 &	 22 &	 6.8 &	 8.2 &	 4.0 &	 52 &	 5.3 \\

DINO & 11.0 &	 5.6 &	 49 &	 7.4 &	 9.6 &	 8.0 &	 17 &	 8.7 &	 11.0 &	 5.6 &	 49 &	 7.4  \\

Grounding DINO & \B 12.9 &	 \B 7.5 &	 42 &	 \B 9.5 &	 \B 10.8 &	 \B 10.0 &	 \B 7 &	 \B 10.4 &	 \B 13.1 &	 \B 7.1 &	 46 &	 \B 9.2   \\

GLIP &  9.8 &	 6.3 &	 \B 36 &	 7.6 &	 8.8 &	 8.2 &	 \B 7 &	 8.5 &	 9.2 &	 5.4 &	 \B 41 &	 6.8  \\
\bottomrule
\end{tabular}%
\caption{Single-source DG analysis of SOTA detectors on RWDS-CZ.}
\label{tab:rwds_cz_ss_results}
\end{table*}
\begin{table*}[t]
\centering
\renewcommand{\arraystretch}{1.1}
\fontsize{9pt}{9pt}\selectfont
\setlength{\tabcolsep}{6pt}
\sisetup{
  table-format=2.1,
  table-number-alignment=center,
  table-text-alignment=center,
  detect-weight        = true,
}
\begin{tabular}{@{}l
                S[table-format=2.1]
                S[table-format=1.1]
                S[table-format=2.0]
                S[table-format=2.1]
                S[table-format=2.1]
                S[table-format=2.1]
                S[table-format=2]
                S[table-format=2.1]
                S[table-format=2.1]
                S[table-format=1.1]
                S[table-format=2]
                S[table-format=2.1]@{}}
\toprule
 & \multicolumn{12}{c}{Target} \\
\cmidrule(lr){2-13}
 & \multicolumn{4}{c}{CZ A} & \multicolumn{4}{c}{CZ B} & \multicolumn{4}{c}{CZ C} \\
 \cmidrule(lr){2-5}\cmidrule(lr){6-9}\cmidrule(lr){10-13}
Methods & {$\text{mAP}_\text{ID}$} & {$\text{mAP}_\text{OOD}$} & {$\text{PD}\downarrow$} & {$\text{H}\uparrow$ } &{$\text{mAP}_\text{ID}$} &{ $\text{mAP}_\text{OOD}$} & {$\text{PD}\downarrow$} & {$\text{H}\uparrow$} & {$\text{mAP}_\text{ID}$} & {$\text{mAP}_\text{OOD}$} & {$\text{PD}\downarrow$} & {$\text{H}\uparrow$} \\
\midrule

Faster R-CNN  & 7.7  &	 4.9 &36 &	 6.0 &	 8.2 &	 7.1 &	 13 &	 7.6 &	 7.7 &	 4.1 &	 47 &	 5.4\\
Mask R-CNN  &   7.5 &	 4.7 &37 &	 5.8 &	 8.1 &	6.9 & 15 &	 7.5 & 7.9 &	 4.3 &	 46 &	 5.6 \\

TOOD  &  8.2 &	 5.0 &	 39 &	 6.2 &	 8.7 &	 7.0 &	 19 &	 7.7 & 8.3 & 4.8 &	 42 &	 6.1  \\

DINO  & 11.6 &	 7.2 &	 38 &	 8.9 &	 11.5 &	 9.6 &	 16 &	 10.4 &	 11.8 &	 7.0 &	 40 &	 8.8  \\

Grounding DINO  &  \B 13.1 &	 \B 8.8 &	 33 & \B 10.5 &	 \B 12.5 &	 \B 11.0 & 12 & \B 11.7&	 \B 13.1 &	 \B 9.3 &	 \B 29 &	 \B 10.9  \\

GLIP &  10.6 &	 8.0 &	 \B 24 &	 9.1 &	9.8 &	 9.2 &	\B 6 &	 9.5 &	 9.8 &	 6.8 &	 31 &	8.0   \\
\bottomrule
\end{tabular}%
\caption{Multi-source DG analysis of SOTA detectors on RWDS-CZ.}
\label{tab:rwds_cz_ms_results}
\vspace{-4mm}
\end{table*}

We conduct all the experiments using the MMDetection toolbox~\cite{mmdetection}. We selected object detectors across classical (Faster R-CNN~\cite{faster_rcnn_paper}, Mask R-CNN~\cite{he2017mask}), recent (DINO~\cite{dino_paper}, TOOD~\cite{tood_paper}), and foundation model-based approaches (Grounding DINO~\cite{grounding_dino_paper}, GLIP~\cite{glip_paper}). Table~\ref{tab:od_arch_bb} presents the top-performing backbone architecture selected for each detector, as evaluated on standard object detection datasets by MMDetection.

To train the object detectors, we perform preprocessing to unify the image sizes of the raw images. We start by cropping all the images into $512\times512$ tiles with an overlapping ratio of 0.2 using SAHI~\cite{sahi} while preserving the original resolution of the images. To ensure a fair comparison of model performances and to mimic real-world conditions where hyperparameter optimisation may be impractical, we use the default hyperparameters specified for each model.

\begin{table*}[t]
\centering
\fontsize{9pt}{9pt}\selectfont
\renewcommand{\arraystretch}{1.1}
\setlength{\tabcolsep}{15pt}
\sisetup{
  table-format=2.1,
  table-number-alignment=center,
  table-text-alignment=center,
  detect-weight        = true,
}
\begin{tabular}{@{}l
                S[table-format=1.1]
                S[table-format=1.1]
                S[table-format=2.0]
                S[table-format=1.1]
                S[table-format=2.1]
                S[table-format=2.1]
                S[table-format=2]
                S[table-format=2.1]@{}}
\toprule
 & \multicolumn{8}{c}{Target} \\
\cmidrule(l){2-9}
 & \multicolumn{4}{c}{India} & \multicolumn{4}{c}{US} \\
 \cmidrule(lr){2-5}\cmidrule(l){6-9}
Methods & \multicolumn{1}{c}{$\text{mAP}_\text{ID}$} & $\text{mAP}_\text{OOD}$ & $\text{PD}\downarrow$ & \multicolumn{1}{c}{$\text{H}\uparrow$ } &\multicolumn{1}{c}{$\text{mAP}_\text{ID}$} & $\text{mAP}_\text{OOD}$ & $\text{PD}\downarrow$ & \multicolumn{1}{c@{}}{$\text{H}\uparrow$ } \\
\midrule
Faster R-CNN  &   4.5  &	  1.3  &	  71  &	  2.0  &	  25.5  &	  1.8  &	  93  & 3.4  \\

Mask R-CNN  &   4.3  &	  1.2  &	  72  &	  1.9  &	  25.9  &	  2.0  &	  92  &	  3.7  \\

TOOD  &    5.1  &	  1.6  &	  69  &	  2.4  &	  27.6  &	  2.4  &	  91  &	  4.4   \\

DINO  &    \B 7.0  &	  2.2  &	  69  &	  3.3  &	  30.8  &	  4.3  &	  86  &	  7.5  \\

Grounding DINO  &   6.7  &	  \B 3.3  &	  \B 51  &	  \B 4.4  &	  \B 31.3  &	  10.8  &	  65  &	  16.1   \\

GLIP &   6.7  &	  \B 3.3  &	  \B 51  &	  \B 4.4  &	  30.7  &	  \B 11.9  &	  \B 61  &	  \B 17.2    \\
\bottomrule
\end{tabular}%
\caption{Single-source DG analysis of SOTA detectors on RWDS-FR.}
\label{tab:rwds_fr_ss_results}
\vspace{-2mm}
\end{table*}

\begin{table*}[t]
\centering
\renewcommand{\arraystretch}{1.1}
\fontsize{9pt}{9pt}\selectfont
\setlength{\tabcolsep}{1.5pt}
\sisetup{
  table-format=2.1,
  table-number-alignment=center,
  table-text-alignment=center, 
  detect-weight        = true,
}
\begin{tabular}{l
                S[table-format=2.1]
                S[table-format=2.1]
                S[table-format=2.0]
                S[table-format=2.1]
                S[table-format=2.1]
                S[table-format=2.1]
                S[table-format=2]
                S[table-format=2.1]
                S[table-format=2.1]
                S[table-format=1.1]
                S[table-format=2]
                S[table-format=2.1]
                S[table-format=1.1]
                S[table-format=1.1]
                S[table-format=2]
                S[table-format=1.1]}
\toprule
 & \multicolumn{16}{c}{Target} \\
\cmidrule(l){2-17}
 & \multicolumn{4}{c}{Florence} & \multicolumn{4}{c}{Michael} & \multicolumn{4}{c}{Harvey} & \multicolumn{4}{c}{Matthew} \\
 \cmidrule(lr){2-5}\cmidrule(lr){6-9}\cmidrule(lr){10-13}\cmidrule(l){14-17}
Methods & \multicolumn{1}{c}{$\text{mAP}_\text{ID}$} & $\text{mAP}_\text{OOD}$ & $\text{PD}\downarrow$ & \multicolumn{1}{c}{$\text{H}\uparrow$ } &\multicolumn{1}{c}{$\text{mAP}_\text{ID}$} & $\text{mAP}_\text{OOD}$ & $\text{PD}\downarrow$ & \multicolumn{1}{c}{$\text{H}\uparrow$ } & \multicolumn{1}{c}{$\text{mAP}_\text{ID}$} & $\text{mAP}_\text{OOD}$ & $\text{PD}\downarrow$ & \multicolumn{1}{c}{$\text{H}\uparrow$ } & \multicolumn{1}{c}{$\text{mAP}_\text{ID}$} & $\text{mAP}_\text{OOD}$ & $\text{PD}\downarrow$ & \multicolumn{1}{c@{}}{$\text{H}\uparrow$ } \\
\midrule
Faster R-CNN  & 34.5 &	 8.6 &	 75 &	 13.8 &	 18.6 &	 6.5 &	 65 &	 9.7 &	 25.1 &	 3.7 &	 85 &	 6.4 &	 1.5 &	 0.3 &	 78 &	 0.5  \\

Mask R-CNN  &   34.0 &	 8.3 &	 76 &	 13.3 &	 19.1 &	 6.9 &	 64 &	 10.1 &	 25.6 &	 3.7 &	 86 &	 6.4 &	 1.7 &	 0.4 &	 78 &	 0.6 \\

TOOD  &  35.7 &	 10.4 &	 71 &	 16.1 &	 21.0 &	 7.1 &	 66 &	 10.6 &	 27.5 &	 4.4 &	 84 &	 7.5 &	 2.4 &	 0.5 &	 78 &	 0.9  \\

DINO  &  36.5 &	 12.0 &	 67 &	 18.0 &	 20.6 &	 7.6 &	 63 &	 11.1 &	 \B 31.4 &	 4.9 &	 84 &	 8.5 &	 2.5 &	 0.8 &	 69 &	 1.2  \\

Grounding DINO  &  39.3 &	 17.4 &	 56 &	 24.2 &	 \B 24.2 &	 9.3 &	 62 &	 13.4 &	 31.0 &	 \B 7.7 &	 \B 75 &	 \B 12.4 &	 3.3 &	 1.2 &	 65 &	 1.7 \\

GLIP &   \B 40.8 &	 \B 19.0 &	 \B 53 &	 \B 25.9 &	 23.9 &	 \B 10.2 &	 \B 57 &	 \B 14.3 &	 29.2 &	 7.0 &	 76 &	 11.3 &	 \B 3.7 &	 \B 1.3 &	 \B 64 &	 \B 2.0 \\
\bottomrule
\end{tabular}%
\caption{Single-source DG analysis of SOTA detectors on RWDS-HE.}
\label{tab:rwds_he_ss_results}
\vspace{-4mm}
\end{table*}
\section{Results and Analyses}\label{sec:Results_and_Analyses}

In the single-source experiment, we evaluate OOD performance by calculating the average performance across models tested on OOD domains, while the ID performance is assessed on the test set of the ID source domain. In contrast, for the multi-source setup, we calculate ID performance as the average performance of all object detectors trained on source domains, while OOD performance is evaluated on the test set of the left-out OOD target domain.

\subsection{RWDS across Climate Zones (RWDS-CZ)}\label{sec:ra_rwds_cz_exp}
\subsubsection{Single-Source DG Experiment}\label{sec:rwds_cz_ss_exp}
In the single-source setup, we train all six detectors on the three climate zones, namely, CZ~A, CZ~B and CZ~C, individually. This results in a total of 18 trained object detectors. We then proceed to evaluate the performance of the trained detectors on the different ID and target OOD test sets, yielding a total of an additional 54 inference experiments. Table~\ref{tab:rwds_cz_ss_results} summarises the performance of the detectors on each climate zone.\footnote{The detectors' cross-domain results on RWDS-CZ under single-source setup are in Supplementary ~\ref{sup_ssec:rwds_cz_exp_ss}. }

When comparing the performance on OOD climate zones to the ID test sets, it can be observed from Table~\ref{tab:rwds_cz_ss_results} that all object detectors exhibit a significant performance drop of above 35\% for CZ~A, 7\% for CZ~B and 40\% for CZ~C, highlighting the challenges posed by domain shift across different climate zones and the limitations of current models in handling OOD data efficiently. 

While GLIP experiences the lowest drop between ID and OOD performance for all the climate zones, Grounding DINO achieves the highest overall tradeoff, balancing both ID and OOD performance most effectively. Moreover, highlighted by H-scores, among the SOTA object detectors evaluated, Grounding DINO outperforms other detectors, both in terms of ID and OOD performance. A plausible explanation to this observation could be that Grounding DINO was designed to generalise to unseen classes in an open-set setting and such capabilities not only boost the performance in an open-set setting but also under a DG setting. The qualitative performance across domains are analysed in Supplementary~\ref{sup_ssec:qualitative_DG_Performance_Comparison}. 

\vspace{-2mm}
\subsubsection{Multi-Source DG Experiment}
Similar to the single-source experiment, we evaluate the performance of the trained detectors on the different ID and OOD test sets, yielding a total of an additional 54 experiments. Table~\ref{tab:rwds_cz_ms_results} summarises the performance of each detector across the various combinations of the climate zones.\footnote{The detectors' cross-domain results on RWDS-CZ under multi-source setup are in Supplementary ~\ref{sup_ssec:rwds_cz_exp_ms}.}

When comparing the performance of the object detectors trained on a single-source to those trained under the multi-source setup, it can be observed that training on multiple source domains enhances not only the object detector's performance on the ID test set but also provides a more substantial boost to the OOD performance. 

Furthermore, consistent with the findings of the single-source experiment, all object detectors evaluated under the multi-source setting experience a notable performance drop when tested on OOD target test set compared to the ID test set. Moreover, across the object detectors benchmarked, Grounding DINO demonstrates the highest performance on the  ID and OOD test sets, as evidenced by its H-score.

\subsection{RWDS across Flooded Regions (RWDS-FR)}\label{sec:ra_rwds_fr_exp}

Since RWDS-FR includes only India and the US as domains, this dataset inherently fits within a single-source setting. We train a total of 12 object detectors and conduct a total of 24 experiments to evaluate their ID and OOD performances, respectively. Table~\ref{tab:rwds_fr_ss_results} summarises the ID and OOD performance of all the object detectors.\footnote{The detectors' cross-domain results on RWDS-FR are in Supplementary~\ref{sup_ssec:rwds_fr_exp_ss}.}

A notable decline in performance can be observed across both India and the US on the OOD test sets with drops exceeding 52\% and 62\%, respectively, indicating a significant impact of the domain-specific variations between the two regions. Grounding DINO and GLIP exhibit comparable performance, both outperforming other object detectors in terms of ID and OOD performance. Their superior OOD performance highlights their slight robustness in handling domain shifts, making them more effective than the other object detection models when the domain shift is defined in terms of disparate geographic regions. We present qualitative performance analyses across domains in Supplementary~\ref{sup_ssec:qualitative_DG_Performance_Comparison_fr}. 

\begin{table*}[t]
\centering
\fontsize{9pt}{9pt}\selectfont
\renewcommand{\arraystretch}{1.1}
\setlength{\tabcolsep}{1.5pt}
\sisetup{
  table-format=2.1,
  table-number-alignment=center,
  table-text-alignment=center,
  detect-weight        = true,
}
\begin{tabular}{l
                S[table-format=2.1]
                S[table-format=2.1]
                S[table-format=2]
                S[table-format=2.1]
                S[table-format=2.1]
                S[table-format=2.1]
                S[table-format=2]
                S[table-format=2.1]
                S[table-format=2.1]
                S[table-format=1.1]
                S[table-format=2]
                S[table-format=2.1]
                S[table-format=1.1]
                S[table-format=1.1]
                S[table-format=2]
                S[table-format=1.1]}
\toprule
 & \multicolumn{16}{c}{Target} \\
\cmidrule(l){2-17}
 & \multicolumn{4}{c}{Florence} & \multicolumn{4}{c}{Michael} & \multicolumn{4}{c}{Harvey} & \multicolumn{4}{c}{Matthew} \\
 \cmidrule(lr){2-5}\cmidrule(lr){6-9}\cmidrule(lr){10-13}\cmidrule(l){14-17}
Methods & \multicolumn{1}{c}{$\text{mAP}_\text{ID}$} & $\text{mAP}_\text{OOD}$ & $\text{PD}\downarrow$ & \multicolumn{1}{c}{$\text{H}\uparrow$ } &\multicolumn{1}{c}{$\text{mAP}_\text{ID}$} & $\text{mAP}_\text{OOD}$ & $\text{PD}\downarrow$ & \multicolumn{1}{c}{$\text{H}\uparrow$ } & \multicolumn{1}{c}{$\text{mAP}_\text{ID}$} & $\text{mAP}_\text{OOD}$ & $\text{PD}\downarrow$ & \multicolumn{1}{c}{$\text{H}\uparrow$ } & \multicolumn{1}{c}{$\text{mAP}_\text{ID}$} & $\text{mAP}_\text{OOD}$ & $\text{PD}\downarrow$ & \multicolumn{1}{c@{}}{$\text{H}\uparrow$ } \\
\midrule
Faster R-CNN  &  32.8 &	 12.7 &	 61 &	 18.3 &	 19.0 &	 8.9 &	 53 &	 12.1 &	 25.0 &	 5.2 &	 79 &	 8.6 &	 1.7 &	 0.4 &	 76 &	 0.6 \\

Mask R-CNN  & 33.6 &	 13.3 &	 60 &	 19.1 &	 19.3 &	 9.1 &	 53 &	 12.4 &	 25.8 &	 5.4 &	 79 &	 8.9 &	 1.6 &	 0.7 &	 56 &	 1.0  \\

TOOD  &  34.2 &	 14.0 &	 59 &	 19.9 &	 19.7 &	 9.6 &	 51 &	 12.9 &	 27.2 &	 5.3 &	 81 &	 8.9 &	 2.2 &	 0.5 &	 77 &	 0.8  \\

DINO  &  37.3 &	 17.0 &	 54 &	 23.3 &	 21.4 &	 10.6 &	 50 &	 14.2 &	 31.3 &	 7.7 &	 75 &	 12.4 &	 2.8 &	 0.8 &	 71 &	 1.2 \\

Grounding DINO  &  39.6 &	 28.2 &	 29 &	 32.9 &	 \B 24.3 &	 \B 12.8 &	 \B 47 &	 \B 16.8 &	 \B 32.2 &	 \B 9.4 &	 \B 71 &	 \B 14.5 &	 3.1 &	 \B 1.5 &	 \B 52 &	 \B 2.0   \\

GLIP &   \B 40.8 &	 \B 30.7 &	 \B 25 &	 \B 35.0 &	 \B 24.3 &	 11.4 &	 53 &	 15.5 &	 30.9 &	 7.8 &	 75 &	 12.5 &	 \B 3.2 &	 1.1 &	 66 &	 1.6  \\
\bottomrule
\end{tabular}%
\caption{Multi-source DG analysis of SOTA detectors on RWDS-HE.}
\label{tab:rwds_he_ms_results}
\vspace{-4mm}
\end{table*}

\subsection{RWDS across Hurricane Events (RWDS-HE)}\label{sec:ra_rwds_he_exp}

\subsubsection{Single-Source DG Experiment}\label{sec:ra_rwds_he_exp_ss}
A significant decline in performance can be observed, in Table~\ref{tab:rwds_he_ss_results}, between hurricane events, suggesting that variations in the nature and characteristics of these events contribute to discrepancies in detection precision.\footnote{The detectors' cross-domain results on RWDS-HE under single-source setup are in Supplementary~\ref{sup_ssec:rwds_he_exp_ss}.} This drop highlights the challenge of generalising disaster object detection across different types of extreme weather events due to domain-specific variations, underscoring the need for further research on developing more robust and generalisable detectors to account for such discrepancies.

Consistent with the RWDS-FR experiment, both Grounding DINO and GLIP achieve the highest performance, with GLIP performing slightly better and excelling in OOD detection compared to the other object detectors. It is also evident that the object detectors have the weakest ID performance when evaluated on the test set of Hurricane Matthew, which suggests that the underlying data is difficult and might suffer from factors such as label noise and class imbalance, as discussed in Section~\ref{sec:D_rwds_HE}, influencing the performance of the models negatively. Qualitative domain performance analyses are provided in Supplementary~\ref{sup_ssec:qualitative_DG_Performance_Comparison_fr}.

\subsubsection{Multi-Source DG Experiment}
In our experimental results illustrated in Table~\ref{tab:rwds_he_ms_results}, we observe a slight performance drop in the ID performance of the object detectors when comparing the multi-source setup to the single-source.\footnote{The detectors' cross-domain results on RWDS-HE under multi-source setup are in Supplementary~\ref{sup_ssec:rwds_he_exp_ms}.} This decline in ID performance suggests that training on multiple source domains introduces additional complexity, which can marginally reduce the model's ability to generalise effectively to the ID domain. Therefore, the inclusion of multiple domains likely causes the model to adapt to a broader set of domain features, which may dilute its focus on optimising the performance specifically on the ID test set. While the multi-source setup generally enhances OOD performance, this comes at the cost of slightly decreased ID accuracy.

However, a notable improvement in OOD performance is observed across nearly all models and domains when compared to the single-source setup. This improvement indicates that the model benefits significantly from exposure to a diverse set of source domains, enhancing its ability to generalise to unseen, OOD domain. Furthermore, consistent with the previous experiments, Grounding DINO and GLIP had a comparably high ID and OOD performances in comparison to the other evaluated object detectors. 

\subsection{Error Analysis of Object Detectors}\label{ssec:error_analysis_tide}
\begin{figure}[t]
\centering
\small
\centerline{\includegraphics[width=0.85\columnwidth,keepaspectratio]{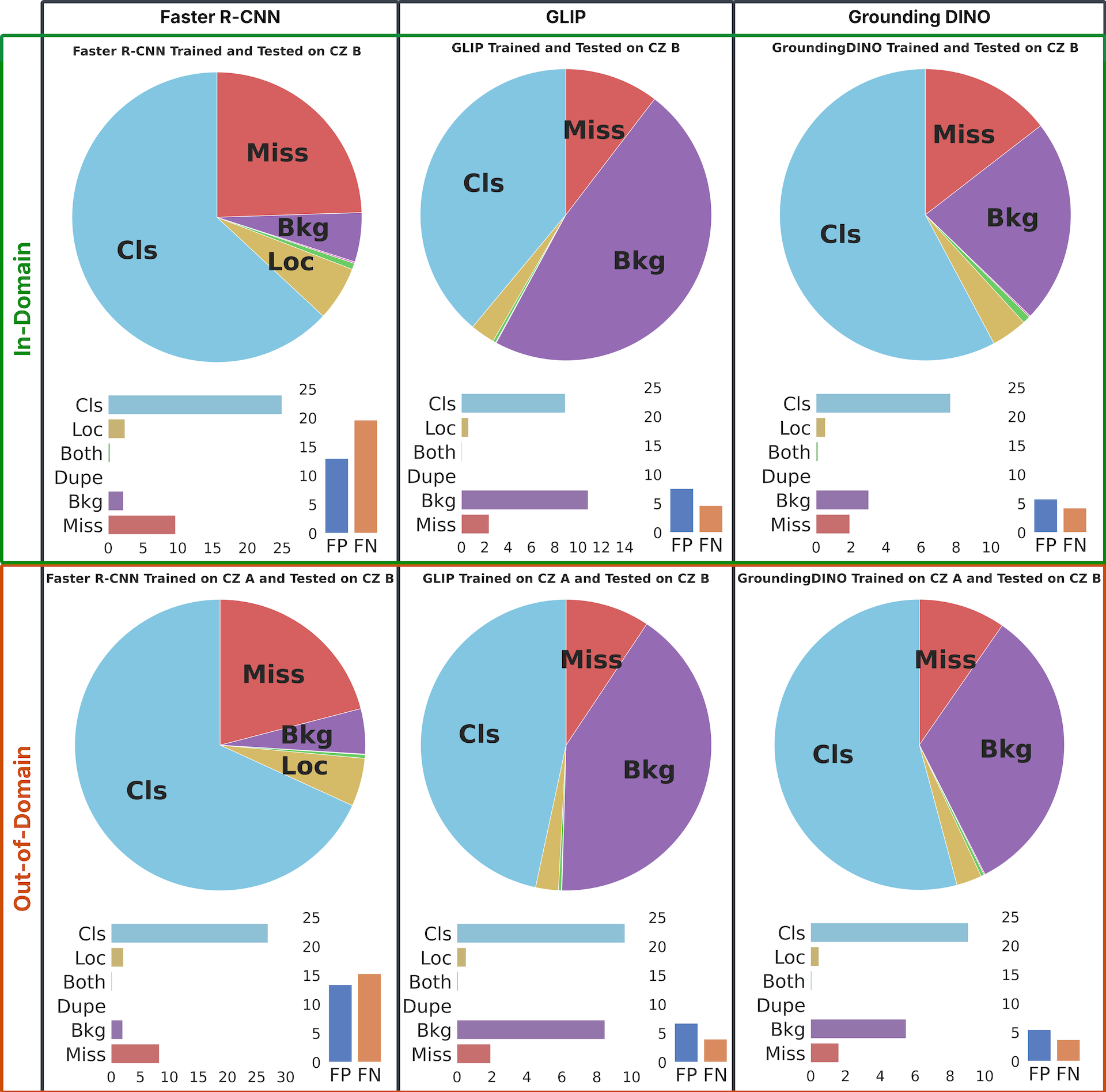}}
\caption{Object detection errors of detectors trained on CZ~A and evaluated on ID (top-row) and OOD (bottom-row) data of CZ~B.}
\label{fig:RWDS_TIDE_Analysis}
\vspace{-3mm}
\end{figure}

We analyse detection errors using the TIDE~\cite{tide} toolbox, as shown in Figures~\ref{fig:RWDS_TIDE_Analysis}. Evaluating on the OOD data of RWDS-CZ as a use-case, where the model is trained on CZ~A and tested on CZ~B, we find that classification errors are the main factor for performance drop, followed by background errors for both GLIP and Grounding DINO, and missed classifications for Faster R-CNN. Moreover, Faster R-CNN has the highest classification and missed groundtruth errors, which aligns with the findings in Section~\ref{sec:ra_rwds_cz_exp}, where it consistently had the weakest performance. While GLIP makes less errors than Faster R-CNN, its higher rate of background errors explains its weaker performance compared to Grounding DINO. A similar trend appears when evaluating the performance of the model trained of CZ~B and evaluated on its ID test set, where Faster R-CNN and Grounding DINO exhibit higher classification errors, while GLIP has the highest background error rate.

\section{Conclusion}\label{sec:Conclusion}
Object detectors typically perform well under the assumption that training and evaluation data come from the same distribution. However, real-world target distributions often differ, causing performance drops due to the distribution shift. DG aims to address this by enabling models to generalise to OOD data without access to target distributions during training. This study examines the generalisability of SOTA object detectors under spatial domain shifts in real world applications and introduces three novel DG benchmark datasets focused on humanitarian and climate change applications. Supported by our findings under single-source and multi-source setups, these datasets, covering domain shifts across climate zones, regions and disaster events, are the first to assess object detection in high-impact real-world contexts and aim to provide valuable resources for evaluating future models' robustness and generalisation.

{
    \small
    \bibliographystyle{ieeenat_fullname}
    \bibliography{main}
}
\clearpage
\setcounter{page}{1}
\setcounter{figure}{0}
\renewcommand{\thefigure}{S\arabic{figure}}

\setcounter{table}{0}
\renewcommand{\thetable}{S\arabic{table}}
\maketitlesupplementary

\appendix

\section{Mathematical Formulation of DG Setups}\label{sup_sec:maths_formulation_dg_setup}
Let $X$ and $Y$ be the input and target spaces, with domain $D$ having joint distribution $P_{XY}$ on $X\times Y$. DG aims to learn a model $f: X \rightarrow Y$ from source data that minimises error on both source (ID) and target (OOD) test data.

\paragraph{\textbf{Single-source domain generalisation.}} We assume that there is only one source domain, $D_{s}$, where $s$ represents a unique source available during the training phase. Therefore, the training set, $D_{train}$, is defined as follows:

\begin{equation}\label{eqn:single_source}
D_{train} =  D_{s} = \{(x_{i}, y_{i})\}_{i=1}^{M}
    \end{equation}
where ${x_i}$, $y_i$ being the $i^{th}$ sample and label pairs from the source domain and $M$ indicating the total number of training samples. Furthermore, $D_{s}$ is associated with a joint distribution $P_{XY}^{s}$. 

\paragraph{\textbf{Multi-source domain generalisation.}} 
We consider a training scenario with access to $N$ distinct yet related source domains, denoted as $D_{s}$ for $s \in \{1, ..., N\}$. Accordingly, the training set is defined as \( D_{\text{train}} \):
\begin{align}\label{eqn:multi-source}
    D_{train} &= \bigcup\limits_{s=1}^{N}D_{s} \\[6pt]
    D_{s} &= \{(x_{i}^{s}, y_{i}^{s})\}_{i=1}^{M_s} \nonumber 
\end{align}
here, $x_{i}^{s}$ represents the $i^{th}$ sample with label $y_{i}^{s}$, and $M_s$ denotes the total number of training samples in domain $D_{s}$. Each source domain $D_{s}$ is characterised by a joint distribution $P_{XY}^{s}$. While the distributions across source domains may be related, they are not equivalent, i.e., $P_{XY}^{s} \neq P_{XY}^{s'}$ for $s \neq s'$, where $s, s' \in \{1, ..., N\}$.

\paragraph{\textbf{The target domain.}} 
We define the OOD target domain(s) as $D_{t}$, where $t$ represents a target domain distinct from the source domains ($t \neq s$). The target domain follows a joint distribution $P_{XY}^{t}$ that differs from all source distributions, i.e., $P_{XY}^{t} \neq P_{XY}^{s}$, $\forall s \in\{1, ..., N\}$. Accordingly, the test set is defined as:
\begin{align}\label{eqn:target-domain}
    D_{test} &= \{D_{t} | t \in \{1, ..., K\}\} \\[6pt]
    D_{t} &= \{(x_{i}^{t}, y_{i}^{t})\}_{i=1}^{M_t} \nonumber 
\end{align}
where $K$ denotes the total number of target domains, $x_{j}^{t}$ is the $j^{th}$ sample with label $y_{j}^{t}$, and $M_t$ signifies the total number of test samples from the target domain $D_t$.

\section{Detailed Benchmarking Results}\label{sup_sec:detailed_benchmarking_results}
In this section, we provide a detailed breakdown of the experiments conducted in our paper, focusing on the individual domains. We begin by analysing the domain shift experienced by the selected object detectors, as discussed in Section~\ref{sec:od_hyp}, across different climate zones using RWDS-CZ in Section~\ref{supp:rwds_cz_exp} under both single- and multi-source setups. Similarly, we discuss the findings related to the generalisation capabilities of the object detectors across flooded regions using RWDS-FR under the single-source setup in Section~\ref{supp:rwds_fr_exp}. Additionally, Section~\ref{supp:rwds_he_exp} presents an examination of the impact of domain shift on the performance of object detectors across various hurricane events in RWDS-HE, also under both single- and multi-source setups.

\begin{table*}[htbp]
\centering
\renewcommand{\arraystretch}{1.1}
\fontsize{8pt}{8pt}\selectfont
\setlength{\tabcolsep}{3pt}
\sisetup{detect-weight        = true,
         tight-spacing        = true,
         table-format         = 2.1,
         table-number-alignment=center,
         table-text-alignment=center
}
\begin{tabular}{@{}llSSSSSSSSSSSS}
\toprule
 & \multicolumn{12}{c}{Target} \\
\cmidrule(lr){3-14}
 & & \multicolumn{4}{c}{CZ~A} & \multicolumn{4}{c}{CZ~B} & \multicolumn{4}{c}{CZ~C} \\
 \cmidrule(lr){3-6}\cmidrule(lr){7-10}\cmidrule(lr){11-14}
Metric & Methods & \multicolumn{1}{c}{$\text{mAP}_\text{ID}$} & $\text{mAP}_\text{OOD}$ & $\text{PD}\downarrow$  & \multicolumn{1}{c}{$\text{H}\uparrow$ } &$\text{mAP}_\text{ID}$ & $\text{mAP}_\text{OOD}$ & $\text{PD}\downarrow$  & \multicolumn{1}{c}{$\text{H}\uparrow$ } & \multicolumn{1}{c}{$\text{mAP}_\text{ID}$} & $\text{mAP}_\text{OOD}$ & $\text{PD}\downarrow$& \multicolumn{1}{c}{$\text{H}\uparrow$ } \\
\midrule

\multirow{6}{*}{$\text{mAP}_{\text{50}}$} & Faster R-CNN &16.7&	9.3&	45&	11.9&	15.7&	13.0&	17&	14.2&	17.4&	8.6&	51&	11.5\\

& Mask R-CNN &16.9&	9.2&	46&	11.9&16.3&	12.6&	23&	14.2&	17.3&	8.8&	49&	11.7\\

& TOOD &17.1&	9.3&	46&	12.0&	15.4&	12.5&	19&	13.8&	17.3&	9.4&	46&	12.2\\

& DINO &25.2&	13.8&	45&	17.8&	19.2&	16.6&	14&	17.8&	24.3&	14.0&	43&	17.7\\

& Grounding DINO & \B 27.9&	\B 17.0&	39&	\B 21.1&	\B 21.5&	\B 20.1&	\B 7&	\B 20.8&	\B 28.1& \B 16.7&	41&	\B 20.9\\

& GLIP &20.7&	13.4&	\B 35&	16.3&	17.1&	15.8&	8&	16.4&	19.2&	12.1&	\B 37&	14.8\\ \midrule

\multirow{6}{*}{$\text{mAP}_{\text{75}}$} & Faster R-CNN &5.2&	2.5&	52&	3.4&	5.9&	4.5&	24&	5.1&	5.3&	2.0&	63&	2.9\\

& Mask R-CNN &4.9&	2.3&	53&	3.1&	5.9&	4.1&	31&	4.8&	5.7&	2.0&	66&	2.9\\

& TOOD &5.8&	2.6&	55&	3.6&	6.8&	5.1&	26&	5.8&	6.8&	2.7&	61&	3.8\\

& DINO&8.2&	4.0&	52&	5.3&	9.0&	6.9&	23&	7.8&	9.2&	4.0&	57&	5.5\\

& Grounding DINO &\B 11.0&	\B 6.1&	45&	\B 7.8&	\B 10.2&	\B 9.1&	11&	\B 9.6&	\B 11.5&	\B 5.4&	53&	\B 7.3\\

& GLIP &8.0&	5.0& \B 38&	6.1& 8.0& 7.4& \B 7& 7.7& 7.5& 4.1& \B 46&	5.3\\ \midrule

\multirow{6}{*}{$\text{mAP}_{\text{50:95}}$} & Faster R-CNN &  7.2 &	 3.9 &	 47 &	 5.0 &	 7.5 &	 6.0 &	 20 &	 6.7 &	 7.7 &	 3.4 &	 56 &	 4.7 \\

& Mask R-CNN &  7.3 &	 3.7 &	 49 &	 4.9 &	 7.7 &	 5.8 &	 25 &	 6.6 &	 7.8 &	 3.5 &	 55 &	 4.8 \\

& TOOD &  7.8 &	 4.0 &	 49 &	 5.2 &	 7.8 &	 6.1 &	 22 &	 6.8 &	 8.2 &	 4.0 &	 52 &	 5.3 \\

& DINO & 11.0 &	 5.6 &	 49 &	 7.4 &	 9.6 &	 8.0 &	 17 &	 8.7 &	 11.0 &	 5.6 &	 49 &	 7.4  \\

& Grounding DINO & \B 12.9 &	 \B 7.5 &	 42 &	 \B 9.5 &	 \B 10.8 &	 \B 10.0 &	 \B 7 &	 \B 10.4 &	 \B 13.1 &  \B 7.1 &	 46 &	 \B 9.2   \\

& GLIP &  9.8 &	 6.3 &	 \B 36 &	 7.6 &	 8.8 &	 8.2 &	 \B 7 &	 8.5 &	 9.2 &	 5.4 &	 \B 41 &	 6.8  \\ 
\bottomrule
\end{tabular}%
\caption{Single-source DG analysis of SOTA detectors on RWDS-CZ where ID/OOD denotes the mAP scores over different IoUs.}
\label{tab:rwds_cz_ss_results_all_regions}
\end{table*}

\paragraph{The Upper Bound Experiments.} To establish a baseline for evaluating model performance under the best-case scenario—where the i.i.d. assumption holds and the model is only tested on samples from the same underlying distribution seen during training, we present the upper-bound (\textbf{UB}) experimental results for RWDS-CZ, RWDS-FR, and RWDS-HE. More specifically, these experiments represent the oracle setup, in which an object detector is trained on the training set from all domains, including the target domain, and evaluated on the test set of each domain respectively.

\subsection{Further Analyses of RWDS-CZ Experiments}\label{supp:rwds_cz_exp}
In Section~\ref{sec:ra_rwds_cz_exp}, we investigated the performance of the selected SOTA object detectors on RWDS-CZ, showing that there exist a shift in the underlying distribution of data gathered from different climate zone. To gain a better intuition on the relationship between these domains, if any, and the influence of domain shift across the different climate zones, we provide a fine-grained analyses of domain shift under the single- (Section~\ref{sup_ssec:rwds_cz_exp_ss}) and multi-source setups (Section ~\ref{sup_ssec:rwds_cz_exp_ms}) along with a qualitative assessment (Section~\ref{sup_ssec:qualitative_DG_Performance_Comparison}) .

\subsubsection{Single-Source DG Experiment}\label{sup_ssec:rwds_cz_exp_ss}
Table ~\ref{tab:rwds_cz_ss_results_all_regions} presents the results of the single-source experiment using mAPs over different IoU regions, namely, $\text{mAP}_{\text{50}}$, $\text{mAP}_{\text{75}}$ and $\text{mAP}_{\text{50:95}}$. The overall trends discussed in Section~\ref{sec:ra_rwds_cz_exp} remain consistent across evaluations using $\text{mAP}{\text{50}}$, $\text{mAP}{\text{75}}$, and $\text{mAP}_{\text{50:95}}$.

Furthermore, Table~\ref{tab:supp_rwds_cz_exp_ss} illustrates the performance of object detectors on UB, which is underlined in the table, CZ~A, CZ~B and CZ~C for each of the six object detectors under the single-source setup. The diagonal, indicated in bold, highlights their ID performance. Aligned with the observations made in Section~\ref{sec:rwds_cz_ss_exp}, there is always a performance drop when testing the OOD test sets. Furthermore, the performances on the UB is always higher than not only the OOD but also the ID. A plausible explanation is that the models benefits from being exposed to a more diverse data distributions during training which makes them more robust in comparison to training on a single source domain.

\begin{figure*}[htbp]
\centering
\small
\centerline{\includegraphics[width=\linewidth,keepaspectratio]{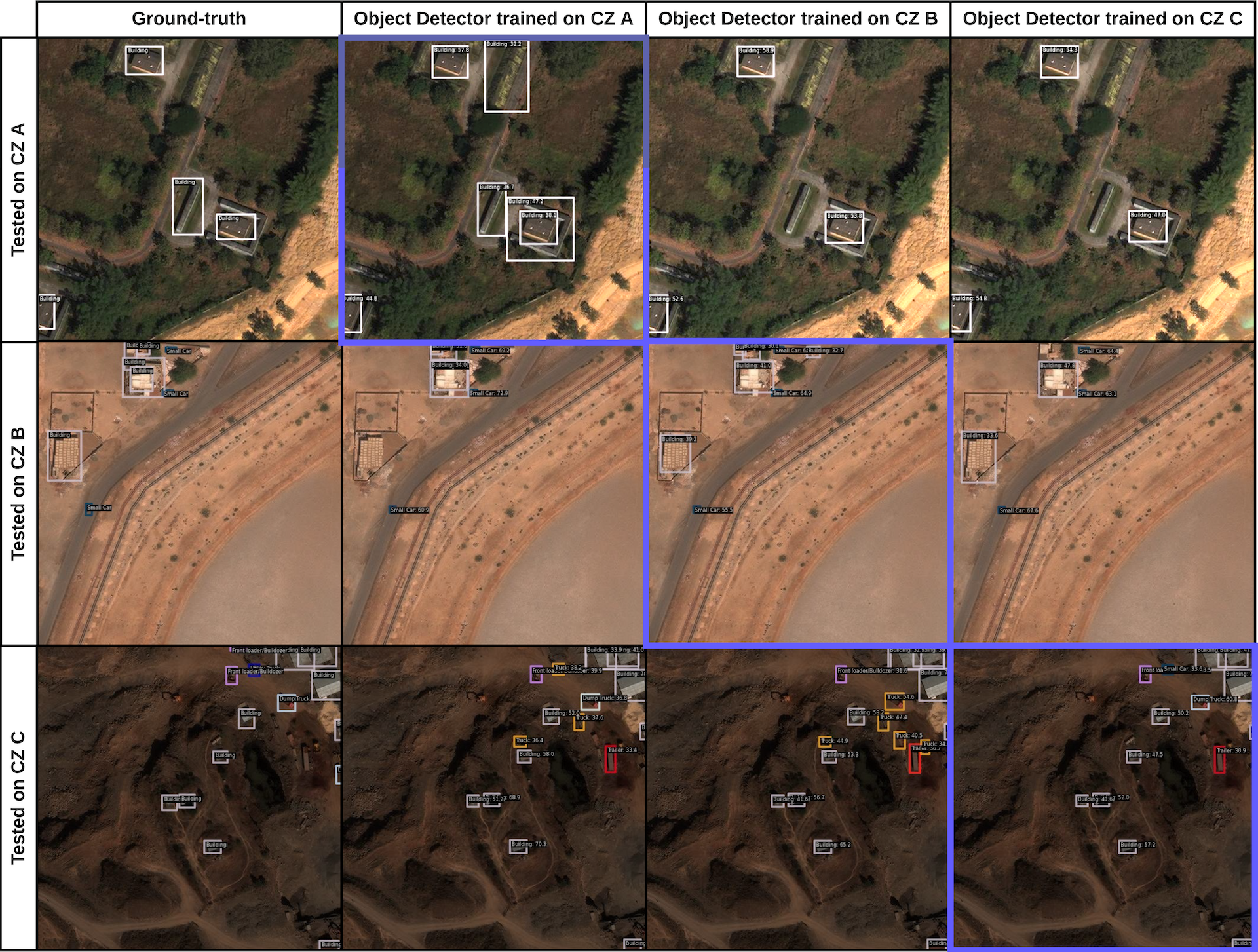}}
\caption{Qualitative DG performance comparison of Grounding DINO among different climate zones, where the diagonal images highlighted in purple indicate the performance on the ID test sample.}
\label{supp_fig:supp_03_Qualitative_Analysis_RWDS_CZ}
\end{figure*}

\begin{table}[htbp]
\centering
\renewcommand{\arraystretch}{1.1}
\fontsize{8pt}{8pt}\selectfont
\setlength{\tabcolsep}{11pt}
\sisetup{detect-weight        = true,
         tight-spacing        = true,
         table-format         = 2.1,
         }
\begin{tabular}{@{}l l S S S@{}}

\toprule
 & & \multicolumn{3}{c}{Target} \\
 \cmidrule(lr){3-5}

Methods & Source   &{CZ~A}& {CZ~B} &{CZ~C} \\ \midrule

\multirow{4}{*}{Faster R-CNN} & \underline{UB}  & \Uline{8.1} &	\Uline{9.0}&	\Uline{7.5} \\
& CZ~A &\B 7.2&	5.3&	3.5  \\
& CZ~B  &3.6&	\B 7.5&	3.3 \\
& CZ~C  & 4.1&	6.7&\B 7.7 \\ \midrule

\multirow{4}{*}{Mask R-CNN} & \underline{UB}  & \Uline{7.8}&	\Uline{9.0}&	\Uline{7.6} \\
& CZ~A & \B 7.3&	5.0&	3.6 \\
& CZ~B  &3.6&	\B 7.7&	3.4\\
& CZ~C  & 3.8&	6.5&	\B 7.8 \\ \midrule

\multirow{4}{*}{TOOD} & \underline{UB}  & \Uline{8.2}&	\Uline{9.1}&	\Uline{8.4}\\
& CZ~A & \B 7.8&	5.0&	4.2  \\
& CZ~B  &3.8&	\B 7.8&	3.7 \\
& CZ~C  & 4.1&	7.1&	\B 8.2 \\ \midrule

\multirow{4}{*}{DINO} & \underline{UB}  &\Uline{12.2}&	\Uline{12.5}&	\Uline{11.8} \\
& CZ~A &  \B 11.0&	7.5&	5.9 \\
& CZ~B  &5.1&	\B 9.6&	5.3\\
& CZ~C  & 6.1&	8.4&	\B 11.0 \\ \midrule

\multirow{4}{*}{Grounding DINO} & \underline{UB}  & \Uline{13.5}&	\Uline{13.8}&	\Uline{12.9}\\
& CZ~A & \B 12.9&	8.6&	7.2  \\
& CZ~B  &6.9&	\B 10.8&	6.9\\
& CZ~C  & 8.1&	11.4&	\B 13.1  \\ \midrule

\multirow{4}{*}{GLIP} & \underline{UB}  & \Uline{11.1}&	\Uline{10.7}&	\Uline{10.1}\\
& CZ~A & \B 9.8&	7.2&	5.7 \\
& CZ~B  &5.8&	\B 8.8&	5.1\\
& CZ~C  & 6.7&	9.2&	\B 9.2 \\ 
\bottomrule

\end{tabular}
\caption{$\text{mAP}_\text{50:95}$ results on RWDS-CZ for the single-source setup.}
\label{tab:supp_rwds_cz_exp_ss}
\end{table}

\subsubsection{Qualitative DG Performance Comparison}\label{sup_ssec:qualitative_DG_Performance_Comparison} 

In order to gain insights on the quality and behaviour of the object detector among the different domains, we select the highest performing object detector, Grounding DINO, and sample the output predictions under the single-source setup. A set of these are illustrated in Figure~\ref{supp_fig:supp_03_Qualitative_Analysis_RWDS_CZ}, where the diagonal images, highlighted in purple, indicate the performance on the ID test samples. It is important to note that we selected the samples with few number of bounding boxes for visualisation purposes.

\begin{itemize}
    \item \textbf{CZ~A}: When tested on CZ~A, aligned with the results found in Table~\ref{tab:supp_rwds_cz_exp_ss}, one can observe that the best performing model in comparison to the ground truth is the one trained on the ID training set. Whereas, the object detectors trained on CZ~B and CZ~C miss detecting a building.

    \item \textbf{CZ~B}: When evaluated on CZ~B, in-line with the results found in Table~\ref{tab:supp_rwds_cz_exp_ss}, the best performing model in comparison to the ground truth is the one trained on the ID training set. Whereas, the object detectors trained on CZ~B and CZ~C miss detecting a number of buildings.

    \item \textbf{CZ~C}: When tested on CZ~C, similar to the findings mentioned in the previous points and the results presented in Table~\ref{tab:supp_rwds_cz_exp_ss}, it can be observed that the best performing model in comparison to the ground truth is the one trained on the ID training set. Whereas, the object detectors trained on CZ~A and CZ~B miss detecting a number of buildings and have higher rates of false negatives and false positives.
\end{itemize}

\begin{table*}[htbp]
\centering
\renewcommand{\arraystretch}{1.1}
\fontsize{8pt}{8pt}\selectfont
\setlength{\tabcolsep}{3pt}
\sisetup{detect-weight        = true,
         tight-spacing        = true,
         table-format         = 2.1,
         table-number-alignment=center,
         table-text-alignment=center
}
\begin{tabular}{@{}llSSSSSSSSSSSS}
\toprule
 & \multicolumn{12}{c}{Target} \\
\cmidrule(lr){3-14}
 & & \multicolumn{4}{c}{CZ~A} & \multicolumn{4}{c}{CZ~B} & \multicolumn{4}{c}{CZ~C} \\
 \cmidrule(lr){3-6}\cmidrule(lr){7-10}\cmidrule(lr){11-14}
Metric & Methods & \multicolumn{1}{c}{$\text{mAP}_\text{ID}$} & $\text{mAP}_\text{OOD}$ & $\text{PD}\downarrow$  & \multicolumn{1}{c}{$\text{H}\uparrow$ } &$\text{mAP}_\text{ID}$ & $\text{mAP}_\text{OOD}$ & $\text{PD}\downarrow$  & \multicolumn{1}{c}{$\text{H}\uparrow$ } & \multicolumn{1}{c}{$\text{mAP}_\text{ID}$} & $\text{mAP}_\text{OOD}$ & $\text{PD}\downarrow$& \multicolumn{1}{c}{$\text{H}\uparrow$ } \\
\midrule

\multirow{6}{*}{$\text{mAP}_{\text{50}}$} & Faster R-CNN &17.5&	11.6&	34&	13.9&	17.0&	14.6&	14&	15.7&	17.5&	10.2&	42&	12.9\\

& Mask R-CNN &  17.2&	11.0&	36&	13.4&	16.7&	14.5&	13&	15.5&	17.9&	10.6&	41&	13.3\\

& TOOD &17.9&	11.3&	37&	13.8&	16.6&	14.3&	14&	15.3&	17.5&	11&	37&	13.5\\

& DINO &25.9&	17.2&	34&	20.7&	22.2&	18.9&	15&	20.4&	25.8&	16.6&	36&	20.2\\

& Grounding DINO &\B 28.9&	\B 19.8&	31&	\B 23.5&	\B 23.9&	\B 21.7&	9&	\B 22.7&	\B 27.7&	\B 21.1&	\B 24&	\B 24.0\\

& GLIP &22.6&	16.7&	\B 26&	19.2&	18.4&	17.4&	\B 5&	17.9&	20.3&	14.6&	28&	17.0\\ \midrule

\multirow{6}{*}{$\text{mAP}_{\text{75}}$} & Faster R-CNN &5.4&	3.4&	37&	4.2&	6.6&	5.4&	18&	5.9&	5.5&	2.3&	58&	3.2\\
& Mask R-CNN &5.2&	3.2&	38&	3.9&	6.6&	5.4&	18&	5.9&	5.8&	2.6&	55&	3.6\\

& TOOD &6.4&	3.6&	43&	4.6&	7.7&	5.8&	25&	6.6&	6.8&	3.4&	50&	4.5\\

& DINO &9.2&	5.4&	41&	6.8&	10.8&	8.8&	18&	9.7&	9.9&	5.5&	44&	7.1\\

& Grounding DINO &\B 10.9&	\B 6.9&	36&	\B 8.4&	\B 12.1&	\B 10.4&	14&	\B 11.2&	\B 11.9&	\B 7.5&	\B 37&	\B 9.2\\

& GLIP &8.6&	6.7&	\B 22&	7.5&	9.2&	8.3&	\B 10&	8.7&	8.3&	5.2&	\B 37&	6.4\\ \midrule

\multirow{6}{*}{$\text{mAP}_{\text{50:95}}$} & Faster R-CNN  & 7.7 &	 4.9 &	 36 &	 6.0 &	 8.2 &	 7.1 &	 13 &	 7.6 &	 7.7 &	 4.1 &	 47 &	 5.4   \\

& Mask R-CNN  &   7.5 &	 4.7 &	 37 &	 5.8 &	 8.1 &	 6.9 &	 15 &	 7.5 &	 7.9 &	 4.3 &	 46 &	 5.6 \\

& TOOD  &  8.2 &	 5.0 &	 39 &	 6.2 &	 8.7 &	 7.0 &	 19 &	 7.7 &	 8.3 &	 4.8 &	 42 &	 6.1  \\

& DINO  & 11.6 &	 7.2 &	 38 &	 8.9 &	 11.5 &	 9.6 &	 16 &	 10.4 &	 11.8 &	 7.0 &	 40 &	 8.8  \\

& Grounding DINO  &  \B 13.1 &	 \B 8.8 &	 33 &	 \B 10.5 &	 \B 12.5 &	 \B 11.0 &	 12 &	 \B 11.7 &	 \B 13.1 &	 \B 9.3 &	 \B 29 &	 \B 10.9  \\

& GLIP &  10.6 &	 8.0 &	 \B 24 &	 9.1 &	 9.8 &	 9.2 &	 \B 6 &	 9.5 &	 9.8 &	 6.8 &	 31 &	 8.0   \\ 

\bottomrule
\end{tabular}%
\caption{Multi-source DG analysis of SOTA detectors on RWDS-CZ where ID/OOD denotes the mAP scores over different IoUs.}
\label{tab:rwds_cz_ms_results_all_regions}

\end{table*}

\subsubsection{Multi-Source DG Experiments}\label{sup_ssec:rwds_cz_exp_ms}
Table ~\ref{tab:rwds_cz_ms_results_all_regions} presents the results of the multi-source experiment using mAPs over different IoU regions, namely, $\text{mAP}_{\text{50}}$, $\text{mAP}_{\text{75}}$ and $\text{mAP}_{\text{50:95}}$, where general trends presented in Section~\ref{sec:ra_rwds_cz_exp} are consistently observed across evaluations utilising $\text{mAP}_{\text{50}}$, $\text{mAP}_{\text{75}}$, and $\text{mAP}_{\text{50:95}}$.

Moreover, Table~\ref{tab:supp_rwds_cz_exp_ms} presents the performance of the six object detectors under the multi-source setup, where an object detector is trained on a collection of source domains and tested on the individual ID test sets in addition to the left out target domain's test set. The diagonal, indicated in bold, highlights their OOD performance.

Unlike the observations made in the single-source setup where the model trained on the UB always had the highest performance, it can be observed from Table~\ref{tab:supp_rwds_cz_exp_ms} that this is not always the case. For example, when trained on the collection of source domains excluding CZ~A, Faster R-CNN, GLIP and Grounding DINO achieve an outstanding performance on the ID test set of CZ~C in comparison to the UB. This suggests that eliminating CZ~A from training actually improves the ID performance of the models on CZ~C. A possible explanation to this phenomena is that the distribution of CZ~A is quite different than that of CZ~B and CZ~C. A similar pattern is observed for multiple other combination of domains and methods, as shown in Table~\ref{tab:supp_rwds_cz_exp_ms}.

\begin{table}[t]
\centering
\renewcommand{\arraystretch}{1.1}
\fontsize{8pt}{8pt}\selectfont
\setlength{\tabcolsep}{7pt}
\sisetup{detect-weight        = true,
         tight-spacing        = true,
         table-format         = 2.1,
         }
\begin{tabular}{@{}l l S S S@{}}
\toprule
 & & \multicolumn{3}{c}{Target} \\
 \cmidrule(lr){3-5}
Methods & Source   & \multicolumn{1}{c}{CZ~A}& \multicolumn{1}{c}{CZ~B} &\multicolumn{1}{c}{CZ~C} \\ \midrule

\multirow{4}{*}{Faster R-CNN} & \underline{UB}  & \Uline{8.1}&	\Uline{9.0}&	\Uline{7.5} \\
& Unseen CZ~A &\B 4.9&	9.1&	7.7\\
& Unseen CZ~B  &7.6&	\B 7.1&	7.7\\
& Unseen CZ~C  & 7.7&	7.3&	\B 4.1\\ \midrule

\multirow{4}{*}{Mask R-CNN} & \underline{UB}  & \Uline{7.8}&	\Uline{9.0}&	\Uline{7.6} \\
& Unseen CZ~A  & \B 4.7&	8.8&	8.0\\
& Unseen CZ~B  & 7.5&	\B 6.9&	7.8\\
& Unseen CZ~C  & 7.4&	7.4&	\B 4.3\\ \midrule

\multirow{4}{*}{TOOD} & \underline{UB}  & \Uline{8.2}&	\Uline{9.1}&	\Uline{8.4}\\
& Unseen CZ~A  & \B 5.0&	9.2&	8.3\\
& Unseen CZ~B  & 8.3&	\B 7.0&	8.3\\
& Unseen CZ~C  &8.0&	8.1&	\B 4.8 \\ \midrule

\multirow{4}{*}{DINO} & \underline{UB}  &\Uline{12.2}&	\Uline{12.5}&	\Uline{11.8} \\
& Unseen CZ~A  & \B 7.2&	11.8&	11.3\\
& Unseen CZ~B  &12.1&	\B 9.6&	12.2 \\
& Unseen CZ~C  & 11.0&	11.1&	\B 7.0\\ \midrule

\multirow{4}{*}{Grounding DINO} & \underline{UB}  & \Uline{13.5}&	\Uline{13.8}&	\Uline{12.9}\\
& Unseen CZ~A  & \B 8.8&	12.9&	13.3\\
& Unseen CZ~B  & 12.8&	\B 11.0&	12.8\\
& Unseen CZ~C  & 13.4&	12.1&	\B 9.3\\ \midrule

\multirow{4}{*}{GLIP} & \underline{UB}  & \Uline{11.1}&	\Uline{10.7}&	\Uline{10.1}\\
& Unseen CZ~A  &\B 8.0&	10.2&	10.3 \\
& Unseen CZ~B  & 10.2&	\B 9.2&	9.3\\
& Unseen CZ~C  & 10.9&	9.4&	\B 6.8\\

\bottomrule
\end{tabular}
\caption{$\text{mAP}_\text{50:95}$ results on RWDS-CZ for the multi-source setup.}
\label{tab:supp_rwds_cz_exp_ms}
\end{table}

\subsection{Further Analyses of RWDS-FR Experiments}\label{supp:rwds_fr_exp}
In Section~\ref{sec:ra_rwds_fr_exp}, we evaluated the performance of the selected object detectors on RWDS-FR, highlighting the existence of distribution shifts in data originating from different flooded regions. To better understand the potential relationships between these domains and the effects of domain shifts across various flooded regions, we provide a detailed analyses of the domain shift under the single-source setup (Section~\ref{sup_ssec:rwds_fr_exp_ss})
accompanied by a qualitative assessment and discussion (Section~\ref{sup_ssec:qualitative_DG_Performance_Comparison_fr}) below.

\subsubsection{Single-Source DG Experiment}\label{sup_ssec:rwds_fr_exp_ss}
As mentioned in Section~\ref{sec:ra_rwds_fr_exp}, RWDS-FR inherently falls under the single-source setup given that it consist of two domains. Table ~\ref{tab:rwds_fr_ss_results_all_regions} presents the results of the single-source experiment using mAPs over different IoU regions, namely, $\text{mAP}_{\text{50}}$, $\text{mAP}_{\text{75}}$ and $\text{mAP}_{\text{50:95}}$. The patterns outlined in Section~\ref{sec:ra_rwds_fr_exp} are observed across evaluations using $\text{mAP}{\text{50}}$, $\text{mAP}{\text{75}}$ (with minor variations), and $\text{mAP}_{\text{50:95}}$.

\begin{table*}[htbp]
\centering
\renewcommand{\arraystretch}{1.1}
\fontsize{8pt}{8pt}\selectfont
\setlength{\tabcolsep}{9.5pt}
\sisetup{detect-weight        = true,
         tight-spacing        = true,
         table-format         = 2.1,
         table-number-alignment=center,
         table-text-alignment=center
}
\begin{tabular}{@{}llSSSSSSSS}
\toprule
 & & \multicolumn{8}{c}{Target} \\
\cmidrule(l){3-10}
& & \multicolumn{4}{c}{India} & \multicolumn{4}{c}{US} \\
 \cmidrule(lr){3-6}\cmidrule(l){7-10}
Metric & Methods & \multicolumn{1}{c}{$\text{mAP}_\text{ID}$} & $\text{mAP}_\text{OOD}$ & $\text{PD}\downarrow$ & \multicolumn{1}{c}{$\text{H}\uparrow$ } &\multicolumn{1}{c}{$\text{mAP}_\text{ID}$} & $\text{mAP}_\text{OOD}$ & $\text{PD}\downarrow$ & \multicolumn{1}{c@{}}{$\text{H}\uparrow$ } \\
\midrule

\multirow{6}{*}{$\text{mAP}_{\text{50}}$} & Faster R-CNN  &  14.7 &	 3.8 &	 74 &	 6.0 &	 56.8 &	 4.6 &	 92 &	 8.5 \\
& Mask R-CNN  &  14.8 &	 3.7 &	 75 &	 5.9 &	 56.7 &	 4.9 &	 91 &	 9.0 \\
& TOOD &  17.2 &	 5.2 &	 70 &	 8.0 &	 59.2 &	 6.0 &	 90 &	 10.9 \\
& DINO &  \B 24.0 &	 8.8 &	 63 &	 12.9 &	 64.6 &	 13.4 &	 79 &	 22.2 \\
& Grounding DINO &  23.3 &	 \B 12.5 &	 \B 46 &	 \B 16.3 &	 \B 67.7 &	 \B 31.3 &	 54 &	 \B 42.8\\
& GLIP &  20.5 &	 11.0 &	 \B 46 &	 14.3 &	 64.0 &	 31.0 &	 \B 52 &	 41.8 \\ \midrule

\multirow{6}{*}{$\text{mAP}_{\text{75}}$} & Faster R-CNN  &  1.7 &	 0.8 &	 53 &	 1.1 &	 19.6 &	 1.1 &	 94 &	 2.1 \\
& Mask R-CNN  &  1.5 &	 0.5 &	 67 &	 0.8 &	 20.2 &	 1.3 &	 94 &	 2.4 \\
& TOOD &  1.5 &	 0.8 &	 \B 47 &	 1.0 &	 22.1 &	 1.5 &	 93 &	 2.8 \\
& DINO &  \B 2.9 &	 0.6 &	 79 &	 1.0 &	 \B 26.7 &	 1.9 &	 93 &	 3.5 \\
& Grounding DINO &  2.6 &	 1.0 &	 62 &	 1.4 &	 25.8 &	 5.1 &	 80 &	 8.5 \\
& GLIP &  \B 2.9 &	 \B 1.1 &	 62 &	 \B 1.6 &	 25.4 &	 \B 6.6 &	 \B 74 &	\B 10.5 \\ \midrule

\multirow{6}{*}{$\text{mAP}_{\text{50:95}}$} & Faster R-CNN  &   4.5  &	  1.3  &	  71  &	  2.0  &	  25.5  &	  1.8  &	  93  & 3.4  \\

& Mask R-CNN  &   4.3  &	  1.2 &	  72  &	  1.9  &	  25.9  &	  2.0  &	  92  &	  3.7  \\

& TOOD  &    5.1  &	  1.6  &	  69  &	  2.4  &	  27.6  &	  2.4  &	  91  &	  4.4   \\

& DINO  &    \B 7.0  &	  2.2  &	  69  &	  3.3  &	  30.8  &	  4.3  &	  86  &	  7.5  \\

& Grounding DINO  &   6.7  &	  \B 3.3  &	  \B 51  &	  \B 4.4  &	  \B 31.3  &	  10.8  &	  65  &	  16.1   \\

& GLIP &   6.7  &	  \B 3.3  &	  \B 51  &	  \B 4.4  &	  30.7  &	  \B 11.9  &	  \B 61  &	  \B 17.2    \\

\bottomrule
\end{tabular}
\caption{Single-source DG analysis of SOTA detectors on RWDS-FR where ID/OOD denotes mAP scores over different IoUs.}
\label{tab:rwds_fr_ss_results_all_regions}
\end{table*}

Furthermore, Table~\ref{tab:supp_rwds_fr_exp_ss} showcases the breakdown of each object detector's performance on the ID and OOD test sets. The bolded diagonal indicates their in-domain performance. While the model trained on India maintains its performance, when comparing the single-source performance versus the UB, the model trained on the US performs slightly better than the UB when evaluated on the ID test set. A plausible explanation for such a behaviour is that the training set of India is naturally difficult and its distribution is further away in the latent space from that of the US, thus hurting the model's ID performance when combined during the training phase. Moreover, aligned with the observations made in Section~\ref{sec:ra_rwds_fr_exp}, the OOD performance of the model trained on India on the US test set is notably low, highlighting the existence of a significant domain shift between the two domains.

\begin{table}[htbp]
\centering
\renewcommand{\arraystretch}{1.1}
\fontsize{8pt}{8pt}\selectfont
\setlength{\tabcolsep}{8pt}
\sisetup{detect-weight        = true,
         table-format         = 2.1,
         table-number-alignment=center,
         table-text-alignment=center
}
\begin{tabular}{@{}llSS@{}}
\toprule
& & \multicolumn{2}{c}{Target} \\
 \cmidrule(lr){3-4}

Methods & Source   &\multicolumn{1}{c}{India}& \multicolumn{1}{c}{United States} \\ \midrule

\multirow{3}{*}{Faster R-CNN} & \underline{UB} &\Uline{4.5}&	\Uline{25.2} \\
& India & \B 4.5&	1.8\\
& United States  & 1.3&	\B 25.5\\ \midrule

\multirow{3}{*}{Mask R-CNN} & \underline{UB}  &\Uline{4.4}&	\Uline{25.8}\\
& India & \B 4.3&	2.0\\
& United States  & 1.2&	\B 25.9\\ \midrule

\multirow{3}{*}{TOOD} & \underline{UB}  &\Uline{5.1}&	\Uline{27.4}\\
& India & \B 5.1&	2.4\\
& United States  & 1.6&	\B 27.6 \\ \midrule

\multirow{3}{*}{DINO} & \underline{UB} &\Uline{7.0}&	\Uline{30.7} \\
& India & \B 7.0&	4.3\\
& United States  & 2.2&	\B 30.8 \\ \midrule

\multirow{3}{*}{Grounding DINO} & \underline{UB}  &\Uline{6.9}&	\Uline{31.2}\\
& India & \B 6.7&	10.8\\
& United States  & 3.3&	\B 31.3 \\ \midrule 

\multirow{3}{*}{GLIP} & \underline{UB}  &\Uline{6.5}&	\Uline{30.8}\\
& India & \B 6.7&	11.9\\
& United States  & 3.3&	\B 30.7 \\ 
\bottomrule

\end{tabular}
\caption{$\text{mAP}_\text{50:95}$ results on RWDS-FR for the single-source setup.}
\label{tab:supp_rwds_fr_exp_ss}
\end{table}

\subsubsection{Qualitative DG Performance Comparison}\label{sup_ssec:qualitative_DG_Performance_Comparison_fr}  
\begin{figure*}[htbp]
\centering
\small
\centerline{\includegraphics[width=0.7\linewidth,keepaspectratio]{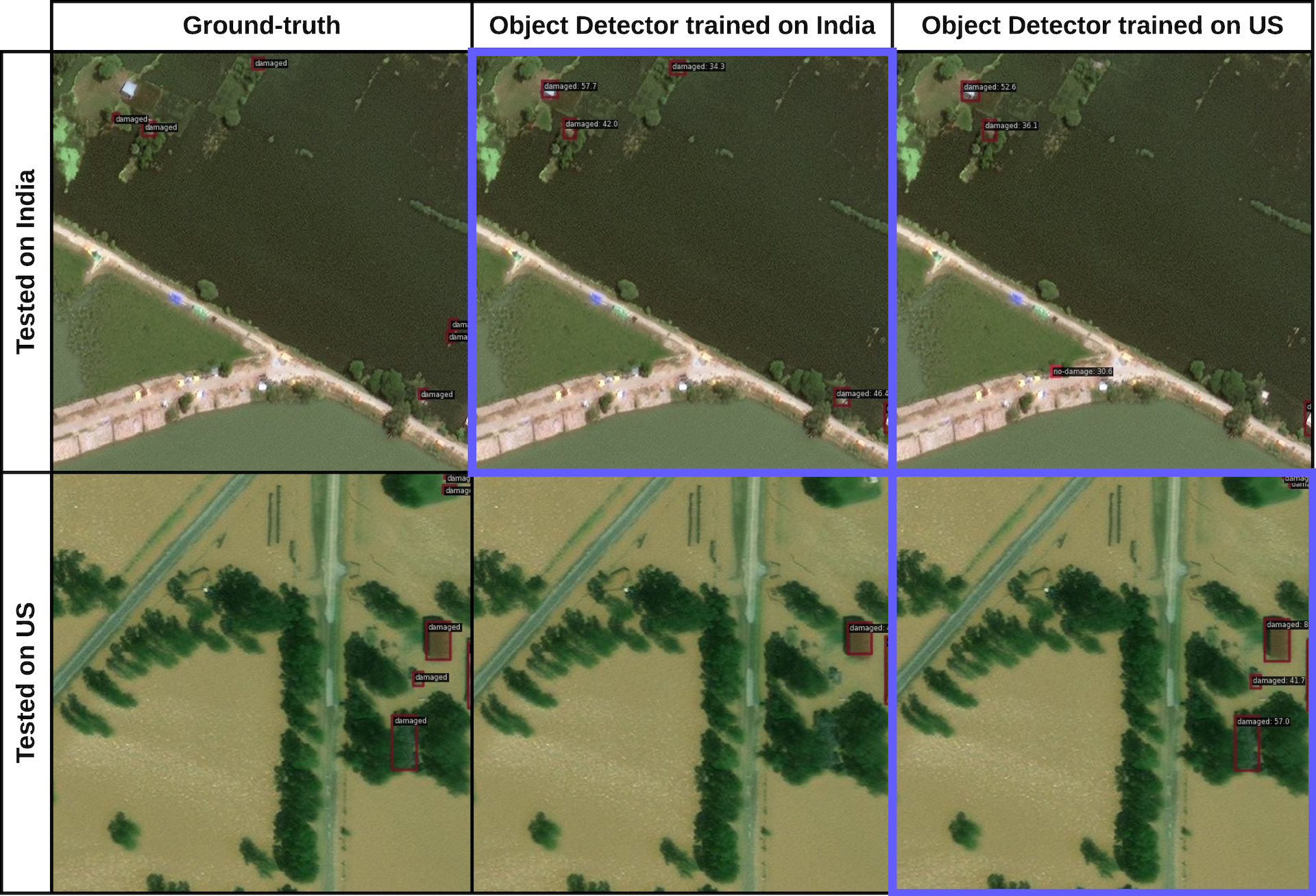}}
\caption{Qualitative DG performance comparison of Grounding DINO among different flooded regions, namely, India and the US, where the diagonal images highlighted in purple indicate the performance on the ID test sample.}
\label{supp_fig:supp_03_Qualitative_Analysis_RWDS_FR}
\end{figure*}

Figure~\ref{supp_fig:supp_03_Qualitative_Analysis_RWDS_FR} illustrates the performance on the ID and OOD test sets of the best performing object detector, Grounding DINO, where the diagonal samples highlighted in purple indicate the performance on the ID test sample. It is worth noting that samples with a limited number of bounding boxes were deliberately chosen to facilitate visualization and enhance clarity in explanation. 

It is evident, from Table~\ref{supp_fig:supp_03_Qualitative_Analysis_RWDS_FR}, that the model trained on India and tested on the ID test set misses a number of bounding boxes. Similarly, the model trained on the US and tested on the OOD test set from India, not only misses a number of bounding boxes, but also consists of false positive detections. However, when evaluated on the test set from the US, its ID performance is closer to the ground-truth. Furthermore, a drop in OOD performance of the model trained on India is observed on when evaluated on OOD US test set, where the model fails in detecting a number of bounding boxes. These observations are aligned with the results previously reported in Table~\ref{tab:supp_rwds_cz_exp_ss}. 

\subsection{Further Analyses of RWDS-HE Experiments}\label{supp:rwds_he_exp}
In Section~\ref{sec:ra_rwds_he_exp}, we analysed the performance of the selected SOTA object detectors on RWDS-HE, emphasising the presence of a distribution shift in data collected from different hurricane events. To gain deeper insights into the potential relationships between these domains and the impact of domain shifts across various hurricane events, we present fine-grained analyses of the object detectors' performance under the single- (Section~\ref{sup_ssec:rwds_he_exp_ss} and multi-source (Section~\ref{sup_ssec:rwds_he_exp_ms}) setups alongside a qualitative assessment (Section~\ref{sup_ssec:qualitative_DG_Performance_Comparison_he}) below.

\subsubsection{Single-Source DG Experiment}\label{sup_ssec:rwds_he_exp_ss}
Table ~\ref{tab:rwds_he_ss_results_all_regions} presents the results of the single-source experiment using mAPs over different IoU regions, namely, $\text{mAP}_{\text{50}}$, $\text{mAP}_{\text{75}}$ and $\text{mAP}_{\text{50:95}}$. The general trends presented in Section~\ref{sec:ra_rwds_he_exp} are consistently observed, with minor variations, across evaluations utilising $\text{mAP}_{\text{50}}$, $\text{mAP}_{\text{75}}$, and $\text{mAP}_{\text{50:95}}$.

\begin{table*}[ht]
\centering
\renewcommand{\arraystretch}{1.1}
\fontsize{8pt}{8pt}\selectfont
\setlength{\tabcolsep}{0.75pt}
\sisetup{detect-weight        = true,
         tight-spacing        = true,
         table-format         = 2.1,
         table-number-alignment=center,
         table-text-alignment=center
}
\begin{tabular}{@{}ll
                S[table-format=2.1]
                S[table-format=2.1]
                S[table-format=2.0]
                S[table-format=2.1]
                S[table-format=2.1]
                S[table-format=2.1]
                S[table-format=2]
                S[table-format=2.1]
                S[table-format=2.1]
                S[table-format=2.1]
                S[table-format=2]
                S[table-format=2.1]
                S[table-format=2.1]
                S[table-format=1.1]
                S[table-format=2]
                S[table-format=1.1]}
\toprule
 & \multicolumn{16}{c}{Target} \\
\cmidrule(l){3-18}
 & & \multicolumn{4}{c}{Florence} & \multicolumn{4}{c}{Michael} & \multicolumn{4}{c}{Harvey} & \multicolumn{4}{c}{Matthew} \\
 \cmidrule(lr){3-6}\cmidrule(lr){7-10}\cmidrule(lr){11-14}\cmidrule(l){15-18}
Metric &Methods & \multicolumn{1}{c}{$\text{mAP}_\text{ID}$} & $\text{mAP}_\text{OOD}$ & $\text{PD}\downarrow$ & \multicolumn{1}{c}{$\text{H}\uparrow$ } &\multicolumn{1}{c}{$\text{mAP}_\text{ID}$} & $\text{mAP}_\text{OOD}$ & $\text{PD}\downarrow$ & \multicolumn{1}{c}{$\text{H}\uparrow$ } & \multicolumn{1}{c}{$\text{mAP}_\text{ID}$} & $\text{mAP}_\text{OOD}$ & $\text{PD}\downarrow$ & \multicolumn{1}{c}{$\text{H}\uparrow$ } & \multicolumn{1}{c}{$\text{mAP}_\text{ID}$} & $\text{mAP}_\text{OOD}$ & $\text{PD}\downarrow$ & $\text{H}\uparrow$ \\
\midrule
\multirow{6}{*}{$\text{mAP}_{\text{50}}$}&Faster R-CNN&64.5&19.3&70&29.7&42.7&17.2&60&24.5&56.9&9.5&83&16.3&5.5&1.2&79&1.9\\
&Mask-CRNN&63.3&18.6&71&28.7&42.9&17.7&59&25.1&57.1&9.5&83&16.2&6.7&1.2&83&2.0\\
&TOOD&64.3&23.2&64&34.1&45.5&18.2&60&26.0&59.9&11.4&81&19.1&7.9&2.1&73&3.3\\
&DINO&66.5&25.9&61&37.2&46.5&19.3&59&27.2&65.8&13.0&80&21.8&9.4&2.8&71&4.3\\
&Grounding DINO&\B 70.6&36.3&49&47.9& \B 52.8&23.8&55&32.8& \B 67.9& \B 20.1& \B 70& \B 31.0& \B 12.5& \B 4.5&64& \B 6.6\\
&GLIP&70.4&\B 37.2& \B 47&\B 48.7&50.8& \B 24.6& \B 52& \B 33.1&63.3&17.5&72&27.5&11.1& \B 4.5& \B 59&6.4\\
 \midrule

\multirow{6}{*}{$\text{mAP}_{\text{75}}$}&Faster R-CNN	&33.2&	6.5&80&10.9&13.8&3.6&74&5.7&19.1&2.1&89&3.8&0.3&0.1&67&0.2\\
&Mask-CRNN&33.7&6.1&82&10.3&14.2&4.1&71&6.4&19.7&2.1&89&3.8&0.4&0.1&75&0.2\\
&TOOD&35.8&7.8&78&12.8&17.0&4.4&74&7.0&22.3&2.7&88&4.8&1.0&0.1&87&0.2\\
&DINO&37.4&9.6&74&15.2&16.3&4.8&71&7.4&\B 27.3&3.0&89&5.4&0.7&0.3&\B 62&0.4\\
&Grounding DINO&40.4&14.5&64&21.3&19.6&5.4&72&8.5&25.2&\B 4.7&81&\B7.9&1.1&0.3&70&0.5\\
&GLIP&\B 42.8& \B 17.2& \B 60& \B 24.6& \B 19.8& \B 6.7& \B 66& \B 10.0&22.9&4.5& \B 80&7.6&\B 1.9& \B 0.4&77& \B 0.7\\
\midrule

\multirow{6}{*}{$\text{mAP}_{\text{50:95}}$} & Faster R-CNN&34.5&8.6&75&13.8&18.6&6.5&65&9.7&25.1&3.7&85&6.4&1.5&0.3&78&0.5\\
&Mask-CRNN&34.0&8.3&76&13.3&19.1&6.9&64&10.1&25.6&3.7&86&6.4&1.7&0.4&78&0.6\\
&TOOD&35.7&10.4&71&16.1&21.0&7.1&66&10.6&27.5&4.4&84&7.5&2.4&0.5&78&0.9\\
&DINO&36.5&12.0&67&18.0&20.6&7.6&63&11.1& \B 31.4&4.9&84&8.5&2.5&0.8&69&1.2\\
&Grounding DINO&39.3&17.4&56&24.2& \B 24.2&9.3&62&13.4&31.0& \B 7.7& \B 75& \B 12.4&3.3&1.2&65&1.7\\
&GLIP&\B 40.8&\B 19.0&\B 53&\B 25.9&23.9&\B 10.2& \B 57& \B 14.3&29.2&7.0&76&11.3& \B 3.7& \B 1.3& \B 64& \B 2.0\\
\bottomrule
\end{tabular}
\caption{Single-source DG analysis of SOTA detectors on RWDS-HE where ID/OOD denotes the mAP scores over different IoUs.}
\label{tab:rwds_he_ss_results_all_regions}
\end{table*}

Furthermore, Table~\ref{supp_tab:rwds_he_exp_ss} outlines the performance of object detectors on UB, Hurricanes Florence, Michael, Harvey and Matthew under the single-source setup. The bolded diagonal indicates their ID performance. In line with the findings in Section~\ref{sec:ra_rwds_he_exp_ss}, all object detectors experience performance degradation when tested on OOD test sets across all domains.

Generally, UB outperforms the other models on the test sets, which is an expected behaviour. However, it can be observed that for rare cases such as when a model, more specifically either of Faster R-CNN, Mask R-CNN or TOOD, is trained on Florence and evaluated on the ID test set, its performance is better than that of the UB. One possible interpretation is that the diversity provided by other distributions hurt the ID performance on Florence compared to training on Florence exclusively.

Furthermore, the results in Table~\ref{supp_tab:rwds_he_exp_ss} clearly show that the model trained on Hurricane Matthew exhibits the weakest performance on both ID and OOD test sets. A likely explanation for this poor performance is that the underlying dataset is challenging and may contain label noise or class imbalance due to the limited number of instances in the raw dataset. These factors, which are independent of domain shift, represent an open area of research and fall outside the scope of this paper.

\begin{table}[t]
\centering
\small

\renewcommand{\arraystretch}{1.1}
\fontsize{8pt}{8pt}\selectfont
\setlength{\tabcolsep}{2.1pt}
\sisetup{detect-weight        = true,
         tight-spacing        = true,
         table-format         = 2.1,
         table-number-alignment=center,
         table-text-alignment=center
}
\begin{tabular}{@{}llSSSS@{}}
\toprule
 & & \multicolumn{4}{c}{Target} \\
 \cmidrule(lr){3-6}

Methods &Source   &\multicolumn{1}{c}{Florence}& \multicolumn{1}{c}{Michael} &\multicolumn{1}{c}{Harvey} &\multicolumn{1}{c}{Matthew}\\ \midrule

\multirow{5}{*}{Faster R-CNN} &\Uline{UB} &\Uline{33.2}&	\Uline{19.3}&	\Uline{25.1}&	\Uline{1.9} \\
&Florence & \B 34.5&	8.4&	5.2&	0.4\\
&Michael  & 8.7&	\B 18.6&	4.8&	0.2\\
&Harvey  & 14.4&	6.9&	\B 25.1&	0.4 \\ 
&Matthew  & 2.8&	4.3&	1.1&	\B 1.5\\ 
\midrule

\multirow{5}{*}{Mask R-CNN} &\Uline{UB} & \Uline{32.8}&	\Uline{19.3}&	\Uline{25.5}&	\Uline{1.7}\\
&Florence & \B 34.1&	9.2&	4.7&	0.4\\
&Michael  & 8.1&	\B 19.1&	4.8&	0.3\\
&Harvey  & 13.5&	7.0&	\B 25.6&	0.4 \\ 
&Matthew  & 3.3&	4.4&	1.5&	\B 1.7\\ \midrule

\multirow{5}{*}{TOOD}&\Uline{UB} &\Uline{34.1}&	\Uline{19.8}&	\Uline{27.7}&	\Uline{2.4} \\
&Florence &	\B 35.7&	9.1&	5.6&	0.5	\\
&Michael &	11.4&	\B 21.0&	6.2&	0.7	\\
&Harvey &	16.1&	7.3&	\B 27.5&	0.4	\\
&Matthew & 	3.7&	5.0&	1.3&	\B 2.4	\\  \midrule

\multirow{5}{*}{DINO}&\Uline{UB} & \Uline{37.7}&	\Uline{22.2}&	\Uline{32.0}&	\Uline{2.8}\\
&Florence &	\B 36.5&	9.6&	6.7&	1.0	\\
&Michael &	11.6&	\B 20.6&	6.3&	0.7	\\
&Harvey &	19.5&	7.8&	\B 31.4&	0.6	\\
&Matthew & 	4.8&	5.4&	1.8&	\B 2.5	\\ \midrule

\multirow{5}{*}{Grounding DINO} & \Uline{UB} &  \Uline{40.4}&	\Uline{24.7}&	\Uline{32.2}&	\Uline{3.0}\\
&Florence &	\B 39.3&	10.5&	8.1&	1.1	\\
&Michael &	18.2&	\B 24.2&	10.2&	1.4	\\
&Harvey &	23.7&	9.4&	\B 31.0&	1.0	\\
&Matthew & 	10.4&	7.9&	4.9&	\B 3.3	\\  \midrule

\multirow{5}{*}{GLIP}&\Uline{UB} & \Uline{41.0}&	\Uline{24.2}&	\Uline{31.1}&	\Uline{3.2}\\
&Florence &	\B 40.8&	10.6&	7.8&	0.9	\\
&Michael &	17.9&	\B 23.9&	9.0&	1.4	\\
&Harvey &	26.3&	10.3&	\B 29.2&	1.7	\\
&Matthew & 	12.8&	9.6&	4.3&	\B 3.7	\\
\bottomrule
\end{tabular}
\caption{$\text{mAP}_\text{50:95}$ results on RWDS-HE for the single-source setup.}
\label{supp_tab:rwds_he_exp_ss}
\end{table}

\subsubsection{Qualitative DG Performance Comparison}\label{sup_ssec:qualitative_DG_Performance_Comparison_he}
\begin{figure*}[htbp]
\centering
\small
\centerline{\includegraphics[width=0.8\linewidth,keepaspectratio]{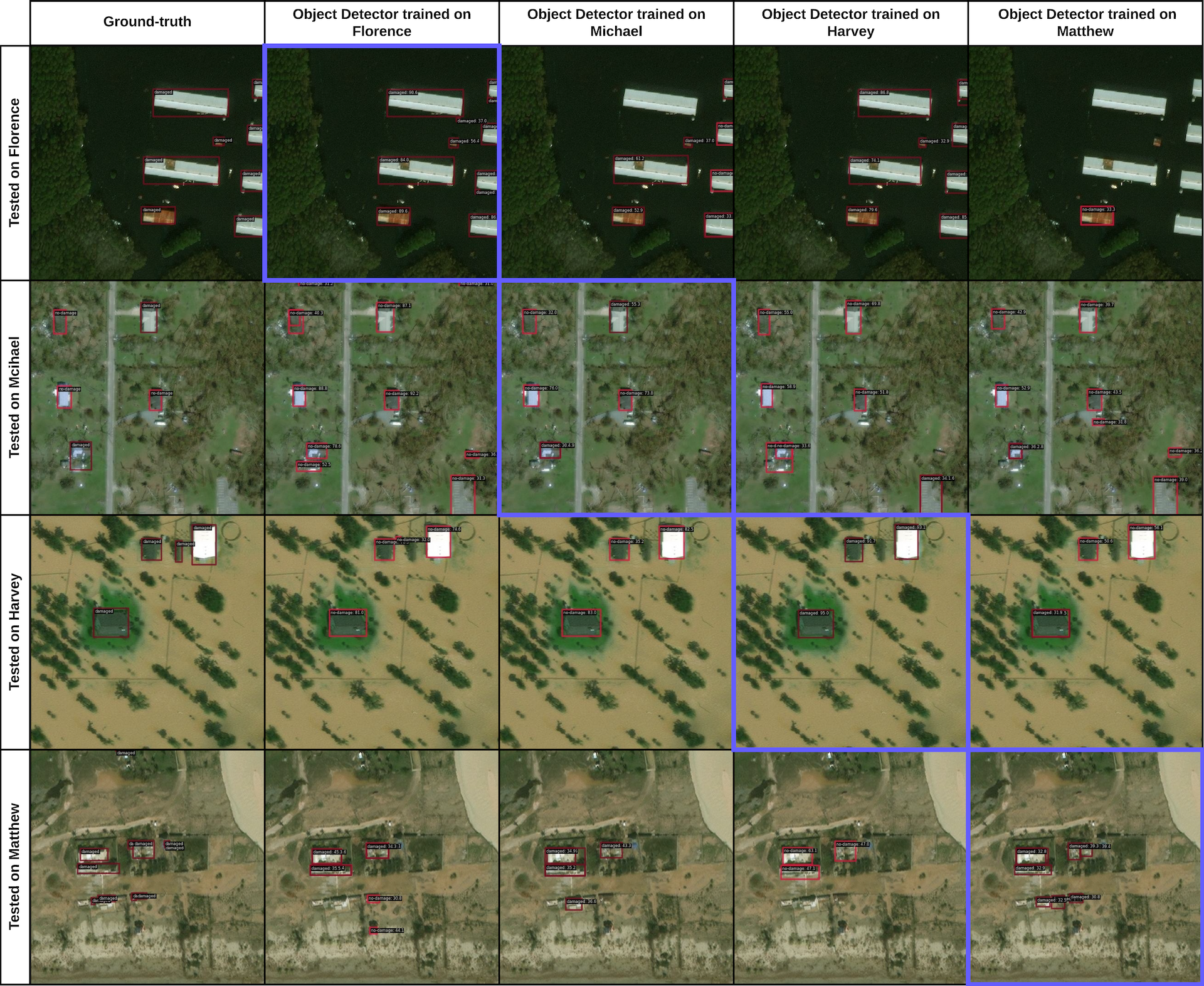}}
\caption{Qualitative DG performance comparison of Grounding DINO among different hurricane events, namely, hurricanes Florence, Michael, Harvey and Matthew, where the diagonal images highlighted in purple indicate the performance on the ID test sample.}
\label{supp_fig:supp_03_Qualitative_Analysis_RWDS_HE}
\end{figure*}

Figure~\ref{supp_fig:supp_03_Qualitative_Analysis_RWDS_HE} illustrates the performance of the best-performing object detector, Grounding DINO, on both ID and OOD test sets. The diagonal samples, highlighted in purple, represent the performance on the ID test samples. Notably, samples with a small number of bounding boxes were intentionally selected to aid visualization and facilitate for clarity in the explanation.

It can be observed that the ID performance across each domain closely matches the ground truth, consistent with the earlier findings from Table~\ref{supp_tab:rwds_he_exp_ss}. However, in certain cases, such as when analysing the ID performance of the model trained on Hurricane Matthew, the model fails to detect several bounding boxes or makes incorrect detections.

Moreover, when examining the OOD performance, the models appear to make similar mistakes during detection. For instance, when testing on the Florence test set, the object detector trained on Florence performs exceptionally well, in contrast to the detectors trained on Michael, where the false negatives are notably higher or in even a worse case, Matthew, where the model fails to generalise effectively.

\subsubsection{Multi-Source DG Experiments}\label{sup_ssec:rwds_he_exp_ms}
Table ~\ref{tab:rwds_he_ms_results_all_regions} presents the results of the multi-source experiment using mAPs over different IoU regions, namely, $\text{mAP}_{\text{50}}$, $\text{mAP}_{\text{75}}$ and $\text{mAP}_{\text{50:95}}$. The results across those three regions exhibit a similar trend to the one reported in Section~\ref{sec:ra_rwds_he_exp}. 

\begin{table*}[htbp]
\centering
\renewcommand{\arraystretch}{1.1}
\fontsize{8pt}{8pt}\selectfont
\setlength{\tabcolsep}{0.75pt}
\sisetup{detect-weight        = true,
         tight-spacing        = true,
         table-format         = 2.1,
         table-number-alignment=center,
         table-text-alignment=center
}
\begin{tabular}{@{}ll
                S[table-format=2.1]
                S[table-format=2.1]
                S[table-format=2.0]
                S[table-format=2.1]
                S[table-format=2.1]
                S[table-format=2.1]
                S[table-format=2]
                S[table-format=2.1]
                S[table-format=2.1]
                S[table-format=2.1]
                S[table-format=2]
                S[table-format=2.1]
                S[table-format=2.1]
                S[table-format=1.1]
                S[table-format=2]
                S[table-format=1.1]}
\toprule
 & \multicolumn{16}{c}{Target} \\
\cmidrule(l){3-18}
 & & \multicolumn{4}{c}{Florence} & \multicolumn{4}{c}{Michael} & \multicolumn{4}{c}{Harvey} & \multicolumn{4}{c}{Matthew} \\
 \cmidrule(lr){3-6}\cmidrule(lr){7-10}\cmidrule(lr){11-14}\cmidrule(l){15-18}
Metric &Methods & \multicolumn{1}{c}{$\text{mAP}_\text{ID}$} & $\text{mAP}_\text{OOD}$ & $\text{PD}\downarrow$ & \multicolumn{1}{c}{$\text{H}\uparrow$ } &\multicolumn{1}{c}{$\text{mAP}_\text{ID}$} & $\text{mAP}_\text{OOD}$ & $\text{PD}\downarrow$ & \multicolumn{1}{c}{$\text{H}\uparrow$ } & \multicolumn{1}{c}{$\text{mAP}_\text{ID}$} & $\text{mAP}_\text{OOD}$ & $\text{PD}\downarrow$ & \multicolumn{1}{c}{$\text{H}\uparrow$ } & \multicolumn{1}{c}{$\text{mAP}_\text{ID}$} & $\text{mAP}_\text{OOD}$ & $\text{PD}\downarrow$ & $\text{H}\uparrow$ \\
\midrule

&\multicolumn{12}{c}{IoU: 0.50} \\ \midrule
\multirow{6}{*}{$\text{mAP}_{\text{50}}$} &Faster R-CNN	&63.1&	27.5&	56&	38.3&	43.3&	20.8&	52&	28.1&	56.6&	13.2&	77&	21.4&	6.0&	1.4&	77&	2.3\\
&Mask-CRNN	&63.8&	28.2&	56&	39.1&	43.9&	21.0&	52&	28.4&	57.8&	14.0&	76&	22.5&	5.8&	2.1&	64&	3.1\\
&TOOD	&62.7&	29.7&	53&	40.3&	43.5&	22.6&	48&	29.8&	59.6&	13.6&	77&	22.1&	7.3&	1.9&	74&	3.0\\
&DINO	&68.2&	35.2&	48&	46.4&	47.3&	24.7&	48&	32.5&	65.9&	18.7&	72&	29.1&	10.6&	3.2&	70&	4.9\\
&Grounding DINO	&70.8&	53.4&	25&	60.9&	\B 52.7&	\B 29.0&	\B 45&	\B 37.4&	\B 69.1&	\B 23.2&	\B 66&	\B 34.7&	\B 11.4&	\B 5.6&	\B 51&	\B 7.5\\
&GLIP	& \B 71.0&	\B 55.8&	\B 21&	\B 62.5&	51.0&	25.2&	51&	33.7&	65.6&	18.4&	72&	28.7&	10.2&	3.9&	62&	5.6\\
 \midrule

\multirow{6}{*}{$\text{mAP}_{\text{75}}$}&Faster R-CNN	&31.4&	9.9&	68&	15.1&	14.5&	6.4&	56&	8.9&	18.6&	3.2&	83&	5.5&	0.5&	0.1&	80&	0.2\\
&Mask-CRNN	&32.2&	11.4&	65&	16.8&	14.6&	6.6&	55&	9.1&	19.6&	3.1&	84&	5.4&	0.4&	0.2&	\B 45&	0.3\\
&TOOD	&33.3&	11.6&	65&	17.2&	15.7&	6.9&	56&	9.6&	21.7&	3.1&	86&	5.4&	0.6&	0.2&	68&	0.3\\
&DINO	&37.1&	14.6&	61&	21.0&	17.3&	7.8&	55&	10.8&	\B 27.5&	5.1&	81&	8.6&	1.0&	0.3&	71&	0.5\\
&Grounding DINO	&40.4&	28.3&	30&	33.3&	19.8&	\B 9.6&	\B 51&	\B 12.9&	26.9&	\B 6.3&	\B 77&	\B 10.2&	1.2&	\B 0.5&	58&	\B 0.7\\
&GLIP	& \B 42.8&	\B 30.7&	\B 28&	\B 35.8&	\B 20.5&	9.0&	56&	12.5&	25.3&	5.4&	79&	8.9&	\B 1.4&	0.4&	71&	0.6\\
\midrule

\multirow{6}{*}{$\text{mAP}_{\text{50:95}}$} &Faster R-CNN	&32.8&	12.7&	61&	18.3&	19.0&	8.9&	53&	12.1&	25.0&	5.2&	79&	8.6&	1.7&	0.4&	76&	0.6\\
&Mask-CRNN	&33.6&	13.3&	60&	19.1&	19.3&	9.1&	53&	12.4&	25.8&	5.4&	79&	8.9&	1.6&	0.7&	56&	1.0\\
&TOOD	&34.2&	14.0&	59&	19.9&	19.7&	9.6&	51&	12.9&	27.2&	5.3&	81&	8.9&	2.2&	0.5&	77&	0.8\\
&DINO	&37.3&	17.0&	54&	23.3&	21.4&	10.6&	50&	14.2&	31.3&	7.7&	75&	12.4&	2.8&	0.8&	71&	1.2\\
&Grounding DINO	&39.6&	28.2&	29&	32.9&	\B 24.3&	\B 12.8&	\B 47&	\B 16.8&	\B 32.2&	\B 9.4&	\B 71&	\B 14.5&	3.1&	\B 1.5&	\B 52&	\B 2.0\\
&GLIP	& \B 40.8&	\B 30.7&	\B 25&	\B 35.0&	\B 24.3&	11.4&	53&	15.5&	30.9&	7.8&	75&	12.5&	\B 3.2&	1.1&	66&	1.6\\

\bottomrule
\end{tabular}
\caption{Multi-source DG analysis of SOTA detectors on RWDS-HE where ID/OOD denotes the mAP scores over different IoUs.}
\label{tab:rwds_he_ms_results_all_regions}
\end{table*}

Moreover, Table~\ref{supp_tab:rwds_he_exp_ms} presents the performance of the object detectors under the multi-source setup, where each detector is trained on a combination of source domains and tested on both the individual ID test sets and the excluded target domain test set. The diagonal, indicated in bold, highlights their OOD performance.

Similar to the observations in the previous section, we can see the domain shift experienced by the object detectors through the performance decline between ID and OOD test sets. Additionally, it is evident that in RWDS-HE, training on multiple domains helps improve the generalisation of the object detectors to OOD test sets, although this may slightly affect the average ID performance due to this trade-off. This is particularly noticeable when examining the OOD performance on Florence for GLIP and Grounding DINO.

\begin{table}[htbp]
\centering
\renewcommand{\arraystretch}{1.1}
\fontsize{8pt}{8pt}\selectfont
\setlength{\tabcolsep}{0.69pt}
\sisetup{detect-weight        = true,
         tight-spacing        = true,
         table-format         = 2.1,
         table-number-alignment=center,
         table-text-alignment=center
}
\begin{tabular}{@{}llSSSS@{}}
\toprule
 & & \multicolumn{4}{c}{Target} \\
 \cmidrule(lr){3-6}

Methods &Source   &\multicolumn{1}{c}{Florence}& \multicolumn{1}{c}{Michael} &\multicolumn{1}{c}{Harvey} &\multicolumn{1}{c}{Matthew}\\ \midrule

\multirow{5}{*}{Faster R-CNN} &\Uline{UB} &\Uline{33.2}&	\Uline{19.3}&	\Uline{25.1}&	\Uline{1.9} \\
&Un.\ Florence & \B 12.7&	18.9&	25.1&	1.9\\
&Un.\ Michael  & 32.4&	\B 8.9&	25.0&	1.5 \\
&Un.\ Harvey  & 32.9&	19.0&	\B 5.2&	1.7 \\ 
&Un.\ Matthew  & 33.1&	19.2&	24.8&	\B 0.4\\ 
\midrule

\multirow{5}{*}{Mask R-CNN}&\Uline{UB} & \Uline{32.8}&	\Uline{19.3}&	\Uline{25.5}&	\Uline{1.7}\\
&Un.\ Florence & \B 13.3&	19.2&	25.4&	1.6\\
&Un.\ Michael  &33.7&	\B 9.1&	25.9&	1.5 \\
&Un.\ Harvey  & 33.5&	19.3&	\B 5.4&	1.7 \\ 
&Un.\ Matthew  & 33.6&	19.4&	26.1&	\B 0.7\\ \midrule

\multirow{5}{*}{TOOD}&\Uline{UB} & \Uline{34.1}&	\Uline{19.8}&	\Uline{27.7}&	\Uline{2.4}\\
&Un.\ Florence &	\B 14.0&	19.7&	27.2&	2.2	\\
&Un.\ Michael &	34.3&	\B 9.6&	27.2&	2.2	\\
&Un.\ Harvey &	34.6&	19.6&	\B 5.3&	2.1	\\
&Un.\ Matthew &	33.6&	19.9&	27.2&	\B 0.5	\\  \midrule

\multirow{5}{*}{DINO}&\Uline{UB} & \Uline{37.7}&	\Uline{22.2}&	\Uline{32.0}&	\Uline{2.8} \\
&Un.\ Florence &	\B 17.0&	21.5&	31.4&	2.7	\\
&Un.\ Michael &	37.6&	\B 10.6&	31.4&	2.9	\\
&Un.\ Harvey &	37.2&	21.2&	\B 7.7&	2.7	\\
&Un.\ Matthew &	37.0&	21.5&	31.1&	\B 0.8	\\  \midrule

\multirow{5}{*}{Grounding DINO}&\Uline{UB} & \Uline{40.4}&	\Uline{24.7}&	\Uline{32.2}&	\Uline{3.0}\\
&Un.\ Florence &	\B 28.2&	24.2&	32.2&	3.0	\\
&Un.\ Michael &	38.4&	\B 12.8&	32.0&	2.8	\\
&Un.\ Harvey &	40.3&	24.3&	\B 9.4&	3.5	\\
&Un.\ Matthew &	40.1&	24.3&	32.3&	\B 1.5	\\  \midrule

\multirow{5}{*}{GLIP} &\Uline{UB} & \Uline{41.0}&	\Uline{24.2}&	\Uline{31.1}&	\Uline{3.2}\\
&Un.\ Florence &	\B 30.7&	24.0&	30.8&	3.5	\\
&Un.\ Michael &	40.1&	\B 11.4&	30.7&	3.1	\\
&Un.\ Harvey &	41.3&	24.5&	\B 7.8&	3.1	\\
&Un.\ Matthew &	40.9&	24.3&	31.3&	\B 1.1	\\
\bottomrule
\multicolumn{6}{l}{\scriptsize * Un.\ means Unseen}
\end{tabular}
\caption{$\text{mAP}_\text{50:95}$ results on RWDS-HE for the multi-source setup.}
\label{supp_tab:rwds_he_exp_ms}
\end{table}
\clearpage

\end{document}